\documentclass[10pt, submission,journal]{IEEEtran}
\usepackage{graphicx}
\usepackage{epsfig}

\usepackage{amssymb}

\usepackage{float}

\usepackage{array}

\usepackage[nodots]{numcompress}

\usepackage{graphicx} 

\usepackage{subfigure}

\usepackage{epsfig} 

\usepackage{amsmath} 

\usepackage{amssymb}  

\usepackage{cite}

\usepackage{url}

\usepackage{float}

\usepackage{graphicx}

\usepackage[misc]{ifsym}

\usepackage{float}
\usepackage{array}
\usepackage[nodots]{numcompress}
\usepackage{graphicx} 
\usepackage{subfigure}
\usepackage{epsfig} 
\usepackage{amsmath} 
\usepackage{amssymb}  
\usepackage{cite}
\usepackage{url}

\usepackage{multirow}
\usepackage{tabularx}
\usepackage{booktabs}
\usepackage{diagbox}
\usepackage[OT1]{fontenc}
\usepackage[normalem]{ulem}

\usepackage[flushleft]{threeparttable}

\usepackage[colorlinks]{hyperref}

\usepackage{xcolor}

\hypersetup{colorlinks=true,citecolor=blue,linkcolor=red,urlcolor=blue}
\usepackage[figuresleft]{rotating}
\newcommand{\tabincell}[2]{\begin{tabular}{@{}#1@{}}#2\end{tabular}}
\newcolumntype{N}{@{}m{0pt}@{}}

\makeatletter
\newcommand{\ssymbol}[1]{\textsuperscript{\@fnsymbol{#1}}}
\makeatother

\begin{document}

\title{Emergent Visual Sensors for Autonomous Vehicles 
}


\author{You Li$^{1}$, \and Julien Moreau$^{2}$, \and
  Javier Ibanez-Guzman$^{1}$ 
  \thanks{$^{1}$You Li and Javier Ibanez-Guzman are with research department of Renault S.A.S, 1 Avenue du Golf, 78280 Guyancourt, France
        {\tt\small \{you.li, javier.ibanez-guzman\}@renault.com}}
\thanks{$^{2}$Julien Moreau is with Universit\'e de technologie de Compi\`egne, CNRS, Heudiasyc (Heuristics and Diagnosis of Complex Systems), CS 60 319 - 60 203 Compi\`egne Cedex, France 
        {\tt\small julien.moreau@hds.utc.fr}}
}




\maketitle

\begin{abstract}
Autonomous vehicles rely on perception systems to understand their surroundings for further navigation missions. Cameras are essential for perception systems due to the advantages of object detection and recognition provided by modern computer vision algorithms, comparing to other sensors, such as LiDARs and radars. However, limited by its inherent imaging principle, a standard RGB camera may perform poorly in a variety of  adverse scenarios, including but not limited to: low illumination, high contrast, bad weather such as fog/rain/snow, etc. Meanwhile, estimating the 3D information from the 2D image detection is generally more difficult when compared to LiDARs or radars. Several new sensing technologies have emerged in recent years to address the limitations of conventional RGB cameras. In this paper, we review the principles of four novel image sensors: infrared cameras, range-gated cameras, polarization cameras, and event cameras. Their comparative advantages, existing or potential applications, and corresponding data processing algorithms are all presented in a systematic manner. We expect that this study will assist practitioners in the autonomous driving society with new perspectives and insights. 

\end{abstract}

\section{Introduction} \label{sec::intro}


Since the dawn of the automotive industry, building a self-driving car has always been a beautiful dream. Owing to the rapid development of sensors, processors and data processing algorithms, developing autonomous vehicles quickly became one of the hottest topics for research and industry, particularly as promoted by the DARPA Grand Challenges from 2004 \cite{Chris2008}. From ADAS (advanced driver assistance systems) to fully autonomous driving (AD), SAE International formally classified the driving automation as 6 levels \cite{SAELevel}. Among those levels, level 2 and 3 are semi-autonomous driving, i.e. the driver is still inside the loop of vehicle control. Level 4 and 5 allow for fully autonomous driving in restricted areas and anywhere, respectively.         

Perception sensors, like human eyes, are critical in scanning and digitizing environments in all levels of autonomous vehicles. The common perception sensors are \textit{visual sensors} like the cameras, and \textit{depth sensors} such as LiDARs \cite{liyouSPM2020}, microwave and ultrasonic radars \cite{radarSPM1027}. By using a \textit{focal plane array} (FPA), a typical camera passively senses the intensities of ambient light at certain wavelengths within its optical \textit{field-of-view} (FOV). Such information is saved as an image, with ambient light intensities sampled as millions of pixel values. A normal camera operates within the visible spectrum that each pixel value is indeed a vector of visible light intensities (e.g.  \textit{red, green}, and \textit{blue}). In this paper, we refer to the camera as a monocular RGB camera by default. LiDARs and radars are sparse active range sensors that measure the distance along with the directions of the transmitted lasers or microwaves. A LiDAR usually has higher accuracy and angular resolution than a radar, whereas a microwave radar can measure velocity using the Doppler effect. In general, cameras mimic human vision and provide rich and dense contextual information. By using range measurements, LiDARs and radars are more accurate than cameras at modeling the 3D world.

The data streams generated by perception sensors are then processed within a perception system to provide useful information for further vehicle navigation. A perception system usually outputs two layers of information: 1) \textit{semantic layer} and 2) \textit{physical layer}. The semantic layer detects and classifies the objects of interest (e.g. pedestrians, vehicles, lane markings, traffic lights, etc), while the physical layer gives them their 3D positions, velocities, and sizes. In general, cameras are superior in the semantic layer, while LiDARs/radars are more reliable in locating the detected objects. Accelerated by the breakthrough of \textit{deep neural networks} (DNN), RGB camera included perception systems have rapidly commercialized and integrated into mass-produced cars in various driving automation levels.  \cite{Brummelen18} and \cite{Yurtsever20} summarize current perception systems and envisage the evolution in the future.      

Despite huge successes, the limitations of RGB cameras in challenging situations have been recognized seriously. Low illumination or other adverse conditions, e.g. fog or rain, can degrade the performance. The glares generated by oncoming headlamps or mirror-like reflections could blind an image. Such image defects would lead to missed detections or unknown behaviors for a perception system, posing safety concerns. To improve the safety and robustness of ADAS/AD, several emerging imaging technologies, e.g. the infrared (IR) cameras, dynamic vision sensors (event cameras), polarization cameras, gated cameras, etc, start to get spotlights. Addressing one or more weaknesses of a conventional RGB camera, those novel image sensors bring extra benefits to complement the common cameras for a better perception system. 

\begin{figure*}[t]
  \centering
   \subfigure[]{
    \includegraphics[width=0.35\textwidth]{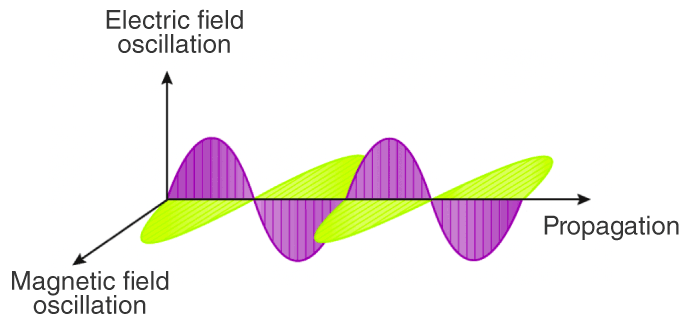}
  }
  \subfigure[]{
    \includegraphics[width=0.5\textwidth]{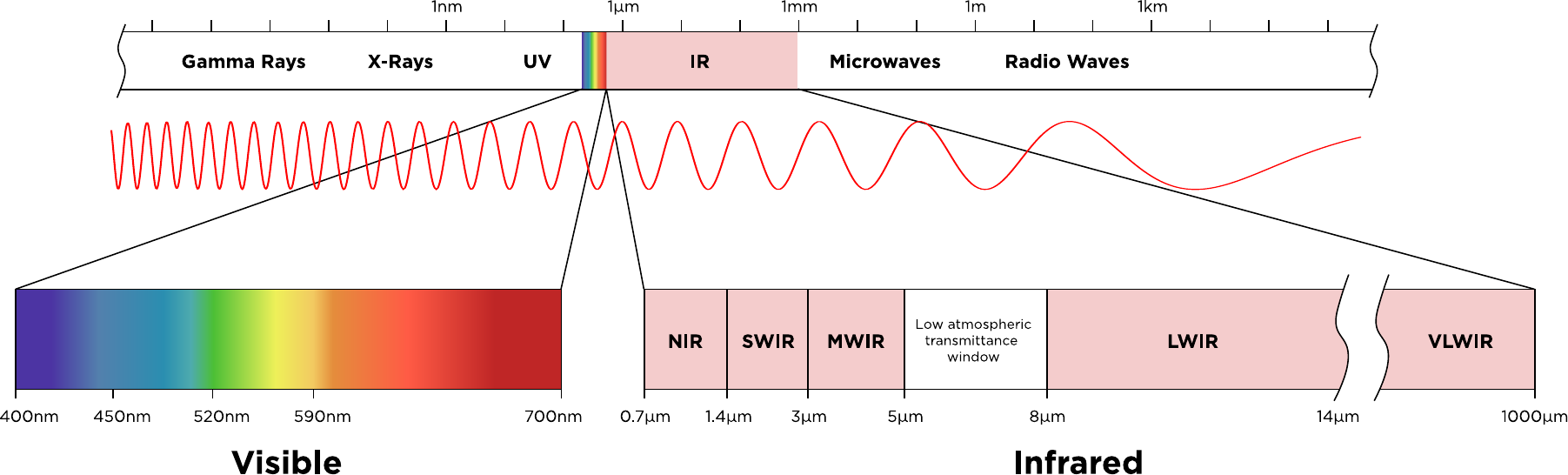}
  }
  \subfigure[ ]{
    \includegraphics[width = 0.45\textwidth]{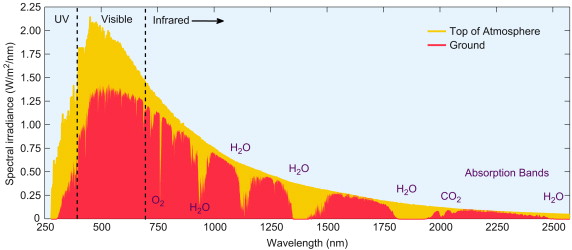}
  }
    \subfigure[]{
    \includegraphics[width = 0.35\textwidth]{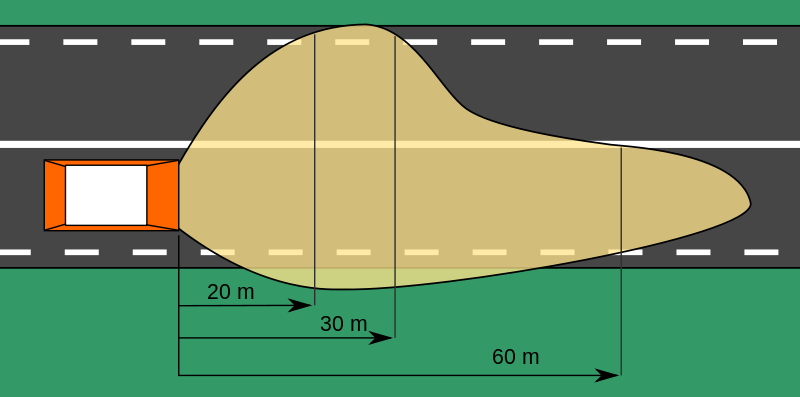}
  }
  \caption{(a) An example of electromagnetic (EM) waves. The mutually perpendicular electric field (in purple) and magnetic field (in purple) are periodically vibrating. (b) Electromagnetic spectrum. Visible light is a special kind of EM waves. (c) Solar irradiance spectrum as a function of wavelength (source from \cite{energy2013}). (d) At night, the area in front of the vehicle is illuminated by the lowbeam headlamps. The maximum range is regulated to be around 60m.}
  \label{fig::IR_spectrum}       
\end{figure*}

To understand the mechanisms of those unconventional cameras, as well as their benefits, potential applications and algorithms, this paper systematically reviews four types of camera: the \textit{infrared camera}, the \textit{polarization camera}, the \textit{range-gated camera} and the \textit{event camera}. This paper is organized as follows: The RGB cameras are reviewed in Sec. \ref{sec::rgb}. Then, the infrared cameras, range-gated cameras, polarization cameras, and event cameras are analyzed in Sec. \ref{sec::IR}, Sec. \ref{sec::gated_camera}, Sec. \ref{sec::polar} and Sec. \ref{sec::event} respectively.

\section{Principle of Conventional RGB camera}\label{sec::rgb}
To understand the limitations of current RGB cameras, we briefly introduce their mechanisms and then analyze the constraints from their imaging principles.

\subsection{The Light} \label{sec::light}
Light is a type of \textit{electromagnetic (EM) waves} that is formed through the interaction between electric and magnetic fields. As shown in Fig. \ref{fig::IR_spectrum} (a), an EM wave is a transverse wave composed of oscillating magnetic and electric fields that are perpendicular to each other, and to the wave's propagation direction as well. Any type of EM wave has three fundamental properties: \textit{amplitude}, \textit{wavelength}, and \textit{polarization}. The wavelength $\lambda$ of visible light ($\lambda \in [400nm, 700nm]$) is only a small portion of the EM spectrum ranging from Gamma rays ($\lambda < 1nm$) to radio waves ($\lambda > 1m$), as shown in Fig. \ref{fig::IR_spectrum} (b). A common RGB camera detects only the intensities i.e. amplitudes of the captured visible light through its lens and is uable to measure polarization information. In a typical road scene, the light is primarily issued from complex interactions between the emitted light from \textit{luminous objects} (e.g. sun, streetlamp, headlamp, etc), the reflected light from \textit{illuminated objects} (e.g. vehicle, pedestrian, building, etc.) and the scattered light from \textit{transmission medium} (e.g. foggy air).           

During the daytime, the sun is the most common source of light. However, human-perceivable sunshine is just a part of the whole solar irradiance on the ground. As shown by Fig. \ref{fig::IR_spectrum} (c), the spectrum of solar irradiance approximately contains 5\% ultraviolet wavelengths, 43\% visible wavelengths, and 52\% infrared wavelengths (values from \cite{ASTMG173}). At night, vehicle's headlamps and streetlamps are the primary source of light \cite{headlamp2003}. However, the lighting pattern of the car headlamps is strictly regulated for safety reasons: the maximum range of low beams can only reach around 60m \cite{lampRegu} (as shown in Fig. \ref{fig::IR_spectrum} (d)), the high beams can reach over 150m but are not allowed to be used continuously. 

The targets of interest (e.g. vehicles, pedestrians, etc) are visible in the images due to the light reflection from their surfaces. Two types of reflection contribute to the imaging results: (1) \textit{diffuse reflection} and (2) \textit{specular reflection}. Rough surfaces, such as asphalt roads and clothing, typically produce diffuse reflections that scatter incident light in various directions. Smooth surfaces, such as metallic material or wet road, would generate specular reflections (a mirror-like reflection) in which the reflected light is concentrated in specific directions determined by the incident angle and the surface property. 

In many cases, the light transmission medium (e.g. air) is assumed to be transparent. However, in adverse conditions, such as fog, rain, snow, or smoke, the floating particles would cause light scattering that results in image blur. Light scattering can be roughly classified as \textit{Mie scattering} or \textit{Rayleigh scattering} based on the particle size to light wavelength ratio. Rayleigh scattering occurs when the particle size is very tiny w.r.t the light wavelength: the blue color of the sky is primarily caused by the Rayleigh scattering of solar irradiance at short wavelengths (e.g. blue at the end of the visible spectrum). For particle sizes similar or larger than a wavelength, such as the water droplet in fog, Mie scattering predominates \cite{duthon2019}.

\subsection{From Light to Digital Images} \label{sec::image_sensor}
The captured light from the various sources is focused by the lens of a camera to its focal plane, where a FPA (i.e. image sensor) is placed to generate images. An image sensor is indeed a 2D array of photosites that can convert light intensities into electrical signals, which are then converted into digits. Each photosite gives a pixel of the image. A photosite is a circuit made up of a \textit{photodetector} and other electronic components. Based on the photoelectric effect of semiconductor material, \textit{Photodiodes} are the most commonly used components. A photodiode \cite{photodiode2009} is a semiconductor that converts light into an electrical signal. When the incident photon energy absorbed by a photodiode exceeds the bandgap of its material, electron-hole pairs (EHPs) are generated. Then, a photocurrent $I_p$ is generated that is approximately linearly proportional to illuminance intensity. A photodiode only responds to specific wavelengths depending on the semiconductor material, which includes, but is not limited to: \textit{silicon (Si)}, \textit{germanium (Ge)}, \textit{indium gallium arsenide (InGaAs)}.  
\begin{figure}[t]
  \centering
\includegraphics[width = 0.3\textwidth]{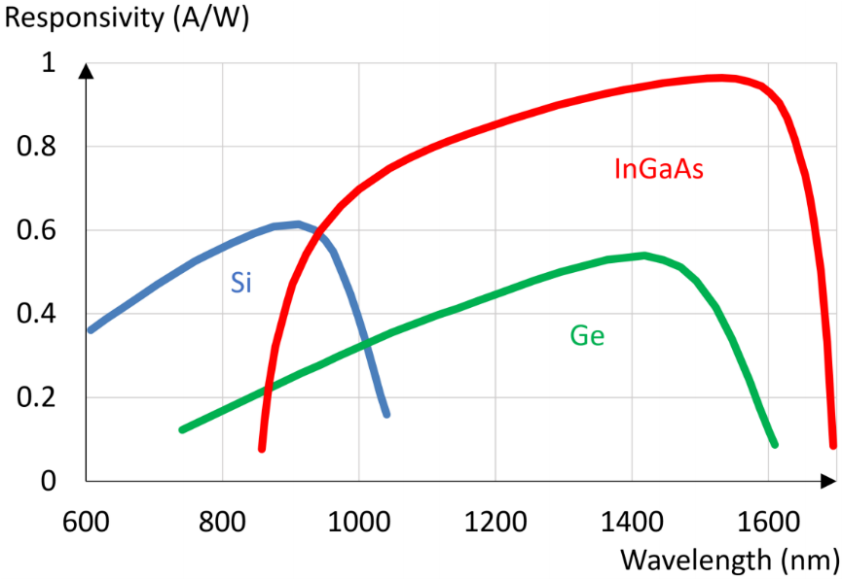}
\caption{ Typical responsivity w.r.t wavelength for Si, InGaAs, and Ge based image sensors (source from \cite{Carrasco-Casado2020}).}
\label{fig::qe_material}       
\end{figure}
Two important metrics represent a photodiode's sensitivity, \textit{quantum efficiency (QE)} and \textit{responsivity}. QE $\eta$ represents the conversion efficiency of photons to electrons. For a specific wavelength $\lambda$, QE $\eta(\lambda)$ is defined as the percentage of photons hitting the photoreactive surface that produces EHPs:  
\begin{equation}
  \eta(\lambda) = \frac{r_e}{r_p} = \frac{\textrm{Electrons Out}}{\textrm{Photons Input}}
\end{equation}

The responsivity $R$ measures the electrical output per optical input. It is defined as the ratio of photocurrent output $I_p$ (in amperes) to the optical power (in watts) $P$:
\begin{equation}
  R(\lambda) = \frac{I_p}{P} = \eta(\lambda)\frac{q}{hf} \approx \eta(\lambda) \frac{\lambda}{1.24} (A/W)
\end{equation}
where $q$ is electron charge, $h$ is Planck's constant and $f$ is the frequency of the optical signal. Fig. \ref{fig::qe_material} shows an example of responsivity curves for three common semiconductor materials, Si, Ge, and InGaAs. Silicon is sensitive to light in the visible and near-infrared spectrum. InGaAs photodiodes can detect wavelengths ranging from 800nm to 2600nm. Connecting a photodiode with resistors and amplifiers creates a photosite that converts the photocurrent into a voltage for further signal processing. An image sensor is created by assembling millions of photosites, together with other components into a 2D array. Currently, the majority of the image sensors are silicon-based and fabricated by CMOS process, and thus are referred to as CMOS Image sensor (CIS).

\begin{figure*}[t]
  \centering
  \subfigure[]{
    \includegraphics[width = 0.8\textwidth]{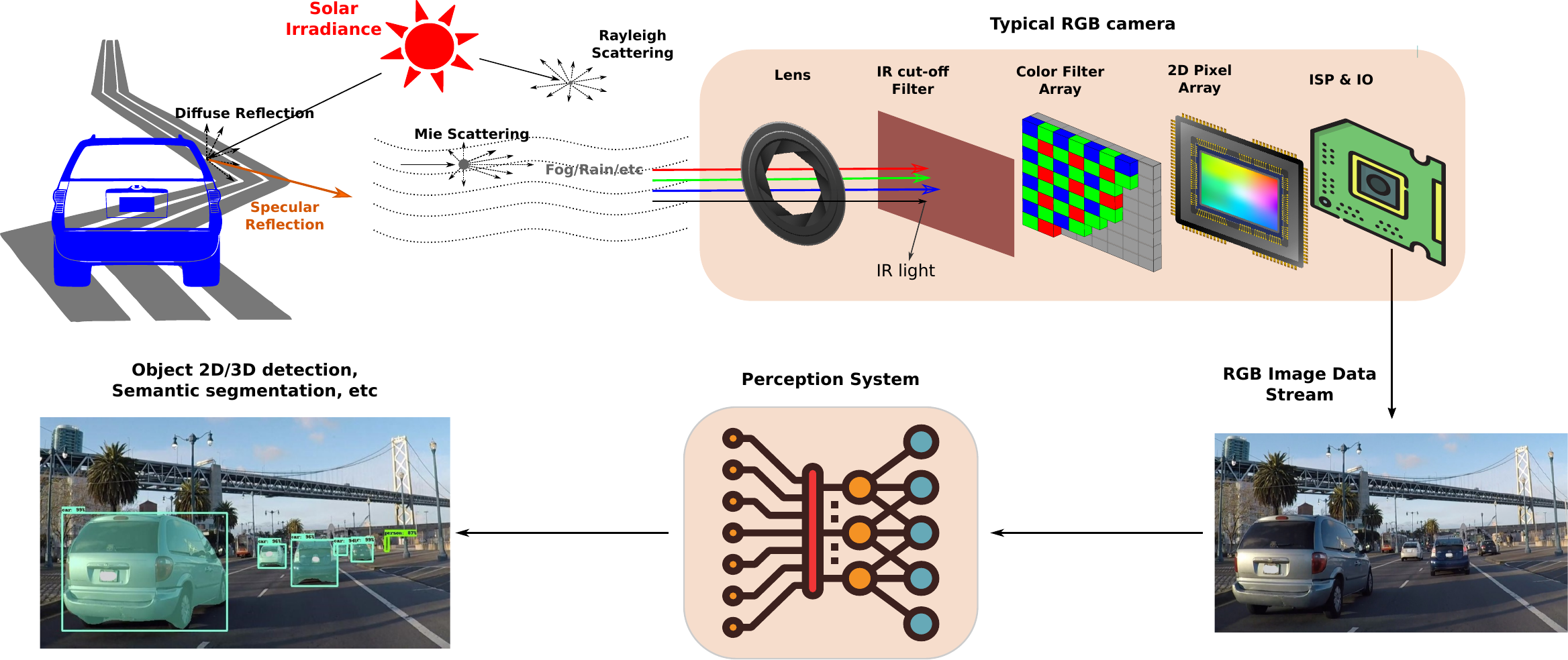}}
  \subfigure[]{
    \includegraphics[width=0.9\textwidth]{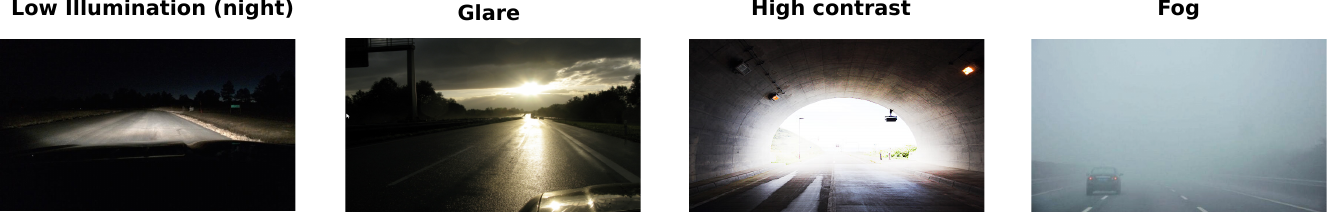}
  }
  \caption{(a) A typical pipeline depicting how the light in a scene is converted to image pixels in a RGB camera before being processed by perception algorithms for environment understanding. The captured light from complex light reflections and scatterings is first passed the IRCF and CFA, allowing the pixels to only respond to red, green, or blue light. Then, within a controlled integration time, electronic signals are generated and converted into digital pixel values inside an image sensor, which are then post-processed in the ISP. Perception algorithms analyze the image outputs for environmental understanding. (b) Four typical difficult scenarios for a common RGB camera (from left to right): low illumination at night, glare caused by specular reflection of wet road, high contrast leads to image over-saturation, a foggy image caused by light scattering.}
\label{fig::camera_pipeline}       
\end{figure*}

By default, image sensors output grayscale values that represent light intensity. To enable color information, a \textit{color filter array (CFA)} is placed just above the image sensor, so that each pixel is sensitive to a specific color wavelength. The \textit{Bayer filter array} is the most common type of CFA, consisting of repeated 2$\times$2 RGGB (red-green-green-blue) filter kernels because the human eye is more sensitive to green light. Only one of the three colors is recorded in the raw output from a Bayer-filter integrated image sensor. In \textit{image signal processor} (ISP), a demosaicing algorithm is implemented to interpolate full color (e.g. a RGB vector) for every pixel. Other types of CFA, such as RCCC or RCCB, are designed to improve traffic light detection or to increase performance under low-light conditions \cite{weikl2020} (with C standing for wideband clear filter, i.e. no color filtering). As shown in Fig. \ref{fig::qe_material}, silicon-based imagers have sensitivities extending into the near-infrared. An \textit{infrared cut-off filter (IRCF)} is designed to block near-infrared wavelengths for better color quality. The entire imaging pipeline is depicted in Fig. \ref{fig::camera_pipeline} (a).

\subsection{Limitations} \label{sec::issues}
Tremendous progress has been made for automotive cameras: the current flagship image sensors, e.g. OmniVision's OX08B24C\footnote{\url{https://www.ovt.com/sensors/OX08B4C}}, or Sony's IMX324\footnote{\url{https://www.sony-semicon.co.jp/products/common/pdf/IMX324\_424.pdf}} can capture ~8MP images at 40fps, and have a 120db dynamic range. However, when confronted with complex road scenarios, several limitations remain, which can be roughly classified as (1) \textit{Image degradation in adverse conditions}, and (2) \textit{Lack depth information}.     

As described in Sec. \ref{sec::light}, a camera is a passive sensor that relies on captured light through a complex interaction between external luminous objects, illuminated targets, and transmission medium. When the received light exceeds the imaging capability, the image quality degrades, affecting the perception results for ADAS/AD. For example, at night, the external illuminations may be insufficient to produce a clear image. During sunny days, specular reflections may appear on the surfaces of the vehicles or the road \cite{roadSpecular2010} that leads to over-saturation. Under adverse weather conditions (e.g. fog, rain, or snow), the strong scattering inside the transmission medium would reduce the image's visibility \cite{duthon2016}. Fig. \ref{fig::camera_pipeline} (b) demonstrates such challenging scenarios for RGB cameras. 

\section{Infrared (IR) Camera} \label{sec::IR}
Conventional RGB cameras only "see" the visible spectrum, as highlighted in Fig. \ref{fig::IR_spectrum} (a). When the light wavelength exceeds 700nm, it enters the "infrared (IR)" spectrum, which is invisible for human and is often divided as follows: (1), \textit{Near-infrared} (NIR): wavelength ranging from 0.7$\mu$m to 1.4$\mu$m. (2), \textit{Short-wavelength infrared} (SWIR): wavelength ranging from 1.4$\mu$m to 3$\mu$m. (3), \textit{Long-wavelength infrared} (LWIR): wavelength ranging from 8$\mu$m to 14$\mu$m. The \textit{Mid-wavelength infrared} is too rare in automotive applications to be included in this paper. The researches and developments of IR cameras for automotive usages mainly focus on NIR, LWIR, and SWIR wavelengths \cite{irIntech17}. NIR and SWIR are "reflected infrared" wavelengths that rely on external light sources such as the sun or other infrared illuminators. NIR and SWIR imagers work similar to RGB imagers in that they directly transform photons to electrical signals. While LWIR is usually referred to as \textit{"thermal infrared"}, a typical LWIR imager converts the thermal radiation to heat, which is then converted to electrical signals. LWIR cameras can image the world solely through thermal emissions and thus do not require any external sources. 

\begin{figure*}[t]
  \centering
  \subfigure[]{
\includegraphics[width = 0.4\textwidth]{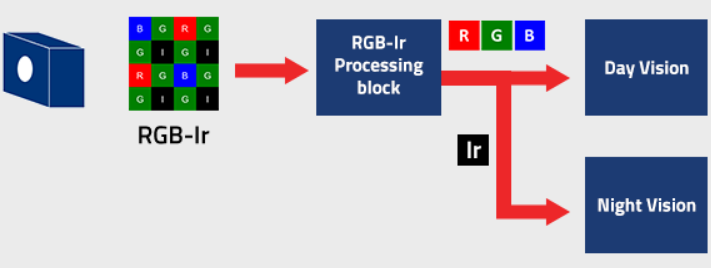}
}
  \subfigure[]{
\includegraphics[width = 0.46\textwidth]{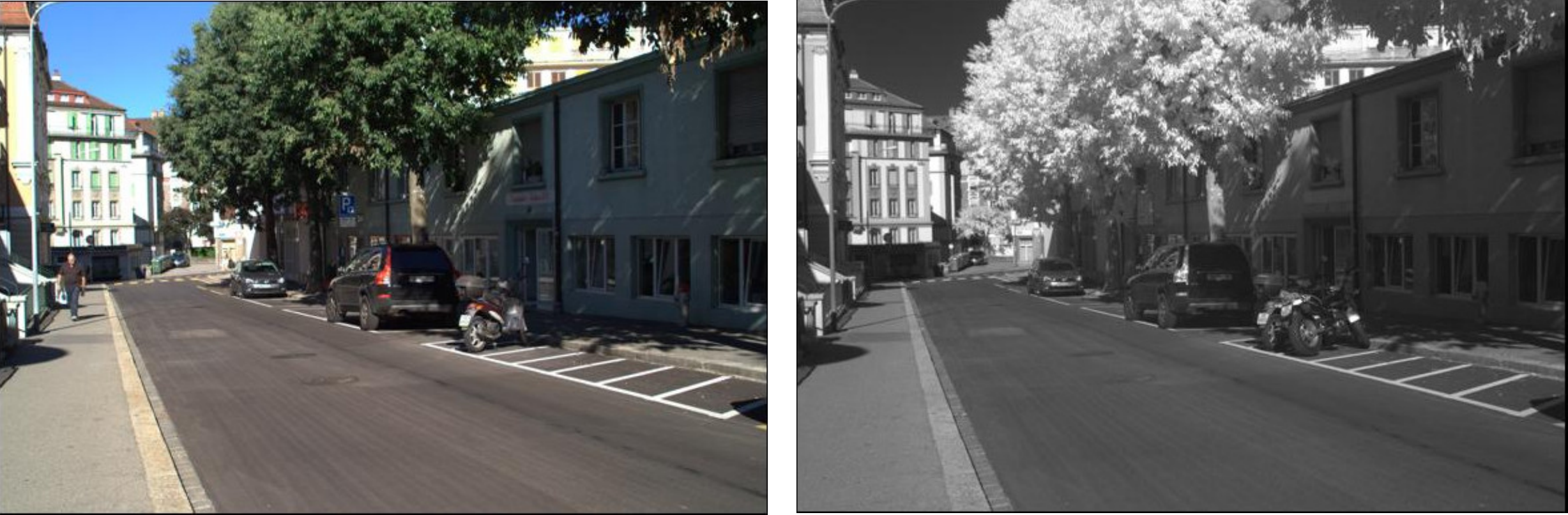}
}
\caption{ (a) A RGB-IR imaging system: the CFA is replaced by a RGB-IR filter array to capture RGB and infrared intensities (from \cite{omni2019}) (b) RGB (left) and NIR (right) images for the same scene. Vegetation in NIR spectrum are "brighter" than in RGB image (from \cite{brownCVPR2011}).}
\label{fig::nir_camera}       
\end{figure*}

\subsection{NIR camera}
NIR imagery shares many properties with RGB imagery: as shown in Fig. \ref{fig::qe_material} (a), a silicon-based imager can still exhibit NIR sensitivity until around 1100nm. As a result, with proper modifications, a RGB camera can be converted into a NIR camera. Because the CFA still has transmission spectra that bleed into NIR wavelengths, removing the IRCF or replacing it with a NIR bandpass filter converts a consumer-grade RGB camera to a NIR camera, as demonstrated in \cite{clement2008} and \cite{jesse2016}. Fig. \ref{fig::nir_camera} (b) shows a RGB image and a NIR image of the same road scene. NIR dedicated pixels are developed to increase the NIR sensitivity. For instance, the Nyxel technology \cite{nyxel2} achieves 50\% QE at 940nm, and 70\% QE at 850nm NIR wavelength. Although other types of materials, such as InGaAs \cite{ingaasNIR2012} may have higher sensitivity in the NIR spectrum, silicon-based image sensors are more popular due to their lower cost.    

In recent years, simultaneously capturing RGB-NIR images has become popular. To achieve that purpose, the CFA, e.g. the Bayer filter, is modified to pass NIR light for specific NIR pixels. Chen \textit{et al.} \cite{chenOE2014} present a four-bandpass filter array to acquire RGB-NIR images. In Lu \textit{et al.}'s work \cite{luyue2016}, a 4$\times$4 pattern containing 15 visible/NIR filters and 1 NIR-only filter, is made. Park \textit{et al.} \cite{parkSensors2016} and Skorka \textit{et al.} \cite{orit2019} further discuss the color distortion and correction problems caused by the RGB-IR filter array. On the industry side, Omnivision has commercialized RGB-NIR imaging systems \cite{omni2019} for automotive applications, as shown in Fig. \ref{fig::nir_camera} (a). A more comprehensive study on RGB-IR camera design could be found in Geelen \textit{et al.}\cite{imec2017}. 

By simply adding external NIR illuminators, usually NIR LEDs (\textit{light-emitting diodes}) or VCSELs (\textit{vertical-cavity surface-emitting lasers}), a passive NIR camera can be converted to an active night vision system. A NIR LED produces a very broad diffused light distribution, whereas a laser produces a narrow beam. For acquiring 2D images, LEDs are more affordable and thus more popular. While VCSELs enable 3D perception applications \cite{roleVCSEL}, e.g. structure light based 3D reconstruction. Two popular wavelengths are 850nm and 940nm. In the early days, 850nm NIR emitters were used because of higher sensitivity than 940nm. However, human eyes can still see a deep red glow from the 850nm emitter in dark conditions. This can be uncomfortable and/or confusing. Currently, 940nm is preferred due to its complete invisibility, and fewer interferences from the natural environment, as solar IR levels at 940nm are less than half compared to 850nm (see Fig. \ref{fig::IR_spectrum} (c)) due to atmospheric absorption.

\subsection{SWIR camera}
Covering the wavelengths ranging from 1.4$\mu$m to 3$\mu$m, the SWIR images are generated by reflected SWIR light like the NIR and RGB cameras. The longer wavelengths of the SWIR spectrum would reduce the scattering effects caused by the small particles existing in the transmission medium. In theory, the SWIR wavelengths can better penetrate fog, smoke, and other adverse weather conditions. At night, the "nightglow" (a night sky radiance emitted from the relaxation of hydroxyl molecules in the atmosphere) comprising mainly SWIR wavelengths ranging from 1.4$\mu$m to 1.8$\mu$m can provide illumination for SWIR cameras \cite{nightglow2013} as well.  

Though silicon-based image sensors have excellent responsivity from visible to NIR spectrum, the bandgap properties of silicon prevent them from having sufficient sensitivity above 1.1$\mu$m. The Indium gallium arsenide (InGaAs) has a lower bandgap, making it the preferred technology for SWIR imaging \cite{ingaasSWIR2013}, as shown in Fig. \ref{fig::qe_material}. In comparison to other semiconductor materials used in the SWIR spectrum e.g. Ge or HgCdTe, InGaAs detectors are cost-effective and high-sensitive while being operated at room temperature \cite{SWIR2008}. However, compared to silicon-based sensors, InGaAs detectors suffer issues of the higher fabrication cost and pixel detects. As a result, some efforts have been made to exploit the potential of silicon-based imagers for SWIR imaging. As in Lv \textit{et al.} \cite{lvCVPR2019}, a deep neural network is trained to approximate the response of an InGaAs sensor and then used to turn a standard silicon-based CMOS sensor into a SWIR image sensor.

\subsection{LWIR (Thermal) camera}
\begin{figure*}[t]
  \centering
  \subfigure[]{
\includegraphics[width = 0.25\textwidth]{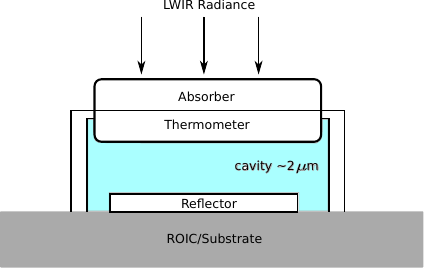}
}
\subfigure[]{
\includegraphics[height = 2.6cm]{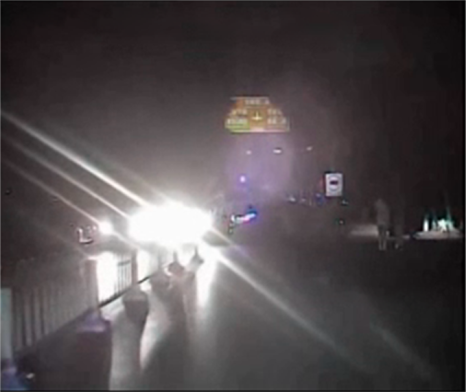}
\includegraphics[height = 2.6cm]{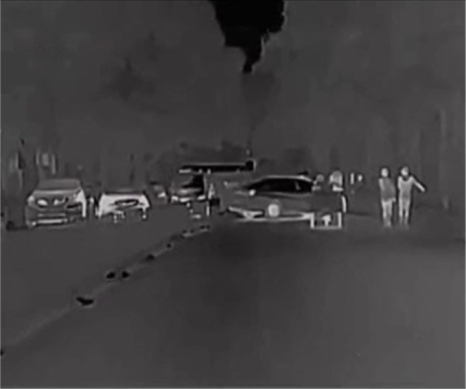}
}
\subfigure[]{
\includegraphics[height = 2.6cm]{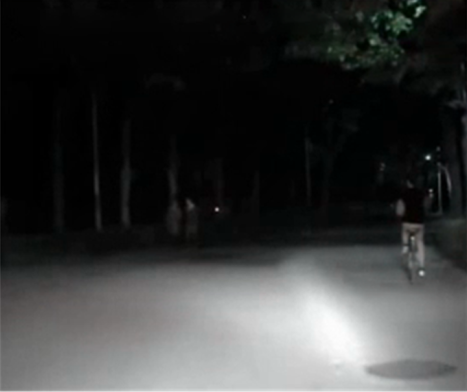}%
\includegraphics[height = 2.6cm]{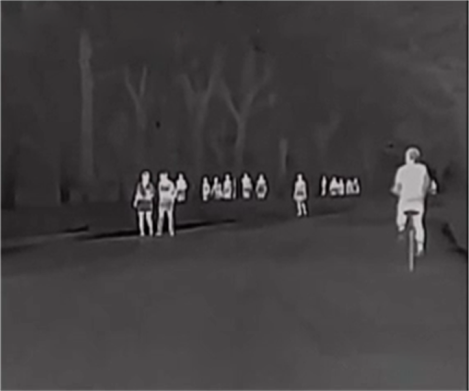}
}
\caption{(a) The architecture of a microbolometer. When exposed to LWIR radiance, the absorber generates heat, which is measured by a thermometer and converted into electrical signals. (b) and (c) RGB images (left) and thermal images (right) of the same scenes (from \cite{dai2020}). The RGB image quality is severely degraded by low illumination, glare, and fog, whereas the thermal camera is unaffected and provides clear images for object detection.}
\label{fig::bolometer}       
\end{figure*}
As a phenomenon of converting thermal energy into electromagnetic energy, all matter with a temperature greater than absolute zero emits thermal radiation. This thermal radiation does not consist of a single wavelength, yet comprises a continuous spectrum. Suppose the radiating matter is ideal, i.e. the black-body, its thermal radiation $B$ for wavelength $\lambda$ is a function of temperature given by Planck's law \cite{planck2006}:  
\begin{equation}
  \centering
  B(\lambda, T) = \frac{2hc^2}{\lambda^5} \frac{1}{e^{hc/ \lambda k_BT}-1}
\end{equation}
where $\lambda, h, c, K_B$ are the wavelength, the Planck's constant, the light speed, and the Boltzmann's constant. Most of the radiation emitted by a human body (\~37°C) is mainly at the wavelength of 12$\mu$m, which is located in the LWIR spectrum. That's the reason for using a LWIR camera for pedestrian/animal detection at night. 

Photon detectors are excellent in thermal imaging because they directly convert the absorbed thermal radiation into electronic changes. However, due to the prohibitively expensive cryogenic cooling systems, their applications in  ordinary scenarios are severely limited. Instead, detecting radiant heat is more popular in LWIR imaging technologies.

Without the need for cooling systems, a \textit{bolometer} \cite{bhan2009} is an instrument that measures heat radiation and converts it into certain measurable quantities. Fig. \ref{fig::bolometer} (a) depicts a block diagram of a bolometer. An absorber and an attached thermometer are deposited above a \textit{read-out integrated circuit} (ROIC) and substrate for the reason of heat insulation. The incident LWIR radiation heats the absorber material, which is typically measured by a thermometer via resistance changes. Historically, the Salisbury screen absorber has been used for bolometers in the LWIR spectrum \cite{LWIRAbsorber}. The \textit{vanadium oxide} (VOx) or the \textit{amorphous silicon} (a-Si) are the common materials for the thermometer layer because they are compatible with standard semiconductor processing technologies, as Tissot \textit{et al.} \cite{ulis2006}, Yon \textit{et al.} \cite{bololeti}, Yu \textit{et al.} \cite{bolonju}. The thermal measures are then transferred to the ROIC for further processing. 

A 2D microbolometer array \cite{microbolo2007} assembled by many tiny uncooled bolometers can capture thermal images. With a much more affordable price and compact size, the microbolometer array is particularly well-suited for mobile applications such as the automobile, especially for the night scenarios Gade and Moeslund \cite{mva2014}. Driven by the rapid progress of semiconductor technologies and MEMS technologies, modern microbolometer arrays can capture images at 60Hz speed with 1024$\times$768 pixels that each pixel is fabricated in 12$\mu$m size. This paper will hereafter refer to thermal or LWIR cameras as an uncooled microbolometer array-based thermal imaging systems.

\subsection{Applications of IR Cameras in Autonomous Vehicles}
Through the introduced principles, the IR cameras can extend a perception system to deal with adverse conditions and with night while avoiding disturbing light emissions for humans. The LWIR camera holds a distinct and unique position because it operates without the need for external light sources. However, on the other hand, since NIR and SWIR cameras are "reflective infrared" like RGB cameras, they provide more context information, e.g. lane markings, tests in traffic signs, etc, than LWIR cameras. Despite some initiatives, SWIR camera applications are uncommon due to the high cost of InGaAs detectors. Here, we mainly review the applications of NIR and LWIR cameras in autonomous vehicles, and automotive night vision system (NVS) is one of the key areas utilizing NIR or LWIR cameras.

Many comparisons and discussions have taken place between the active NIR cameras and passive LWIR night vision systems as in Kallhammer \cite{janerik2006}. Coupled with invisible NIR transmitters (e.g. NIR LEDs or headlamps containing NIR spectrum), an active NIR camera is a cost-effective NVS. While LWIR cameras are particularly suitable for detecting hot-blooded creatures (humans, animals, etc) and other objects with heat signatures (e.g. the engine of a moving vehicle) at night. Fig. \ref{fig::bolometer} (b) (c) depict a comparison of thermal imagery and visible imagery in several harsh conditions. In general, it has been demonstrated in Tsimhoni \textit{et al.} \cite{tsim2004} and \cite{omer2006} that at night, the pedestrian detection range of a LWIR camera (165m) is significantly greater than an active NIR camera (59m). Under other adverse conditions, thermal imaging systems are found to be more stable than NIR cameras. The tests conducted in Judd \textit{et al.} \cite{SPIEflir2019} show that LWIR imaging is significantly less affected by fog than NIR cameras. The experiments conducted by Pinchon \textit{et al.} \cite{whichSpec2018} confirm the advantage of LWIR imagery over NIR imagery in pedestrian detection and demonstrate that the glare caused by oncoming headlamps under fog would not occur in thermal imagery. Nonetheless, the tests in Pinchon \textit{et al.} \cite{whichSpec2018} show that thermal cameras are unable to detect lane markings or recognize traffic signs, whereas NIR imaging systems can. Thermal cameras are thus more effective but limited in detecting pedestrians or animals in various adverse conditions. Because NIR behaves similarly to the visible spectrum, the NIR imagery provides more contextual information and offers more functions. The active NIR cameras are either adopted as cheaper substitutes for thermal imagery in exterior perception systems, or used for in-cabin driver monitoring systems.

In the automotive industry, thermal cameras have surpassed NIR cameras in market share in night vision systems (NVS). In 2000, General Motors launched the first automotive NVS on the Cadillac DeVille using a LWIR sensor supplied from Raytheon \cite{deville2000}. In 2004, Honda \cite{hoda2002} introduced a thermal camera based Intelligent NVS on Honda Legend. From 2005, BMW began to use LWIR cameras in its 7 Series. Peugeut incorporated a thermal camera into its flagship sedan Peugeot 508 from 2018. In 2002, Toyota presented an active NIR NVS in Toyota LandCruiser and Lexus 470, but from 2014 Lexus has decided to discard the NVS in the subsequent generations. When it comes to driver monitoring systems, NIR cameras dominate the market. For instance, in the driver assistant system SuperCruised launched by Cadillac in 2018, an NIR camera is mounted in the instrument panel to monitor whether or not the driver is watching the road in order to raise warnings. In the research field, the situation is similar: most studies on automotive thermal cameras are around pedestrian or animal detection at night. NIR cameras are more investigated for analyzing and monitoring driver's status. Therefore, the following sections focus the in-cabin applications of NIR cameras and pedestrian/animal detections in LWIR imagery.


\begin{figure*}[t]
  \centering
  \subfigure[]{
\includegraphics[width = 0.35\textwidth]{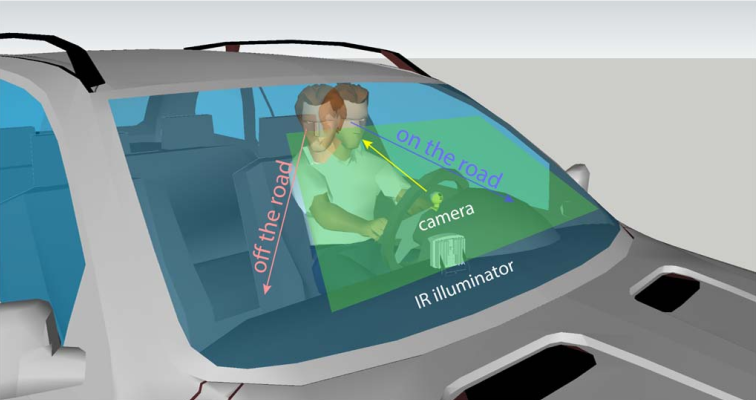}
}
\subfigure[]{
\includegraphics[width = 0.17\textwidth]{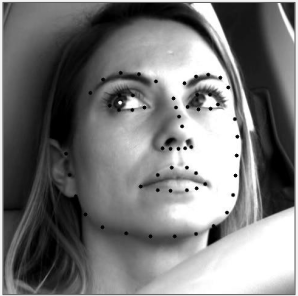}
}
\subfigure[]{
\includegraphics[width = 0.36\textwidth]{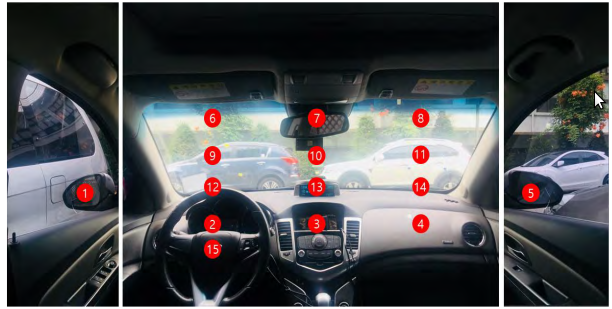}
}
\caption{(a) An example of a DMS consisting of a camera and an IR illuminator facing the driver to detect whether the driver's eyes are on the road or not (from \cite{gazeTracking2015}). (b) Face landmark detection results (from \cite{gazeIET2016}). (c) 15 zones for gaze classification (from \cite{yoon2019}).}
\label{fig::dms}       
\end{figure*}

\subsubsection{NIR Cameras in Driver Monitoring Systems}
According to the NHTSA (National Highway Traffic Safety Administration), approximately 25\% of reported crashes in the U.S. involve a certain form of driver inattention \cite{nhtsa2007}. \textit{Distraction} and \textit{fatigue} are the two principal causes of driver inattention. A visual distraction, such as looking away from the front road, is the most common type of distraction. Fatigue can be defined as a subjective feeling of drowsiness caused by physical or mental factors. A \textit{driver monitoring system} (DMS) utilizes sensors (e.g. image sensor, pressure sensor, etc) to ensure a driver keeping attention on the road, as shown in Fig. \ref{fig::dms} (a). A typical DMS usually contains \textit{gaze detection} and \textit{drowsiness detection} to warn the driver when an inattention event is detected. Researches (Ahlstrom \textit{et al.} \cite{gazeITS2013}, \textit{Schwarz et al.} \cite{dms2019}) have proved that the DMS could effectively improve safety. In Europe, a general safety regulation\footnote{\url{https://ec.europa.eu/commission/presscorner/detail/en/IP_19_1793}} has been passed in 2019 to mandate automakers to install advanced safety systems including DMS in new cars on the EU market from 2022. Because an active NIR night vision system is barely perceptible by human eyes and conserves abundant contextual details, it plays a critical role in modern DMS. Face and eye detectionand tracking via image processing are usually required as a preliminary step before detecting gaze and drowsiness. Fig. \ref{fig::dms} (b) shows an example of detected facial landmarks. In recent years, deep neural networks (DNNs) dominate this domain. For example, Yoon \textit{et al.} \cite{yoon2019} utilize a VGG network for face detection and Park \textit{et al.} \cite{park2019} develop a faster-RCNN based eye detection method.


Following the localization of the face and eye regions, additional processing is required to detect drowsiness or distraction. \textit{PERCLOS (percentage of eye closure over time)} proposed by Dinges \textit{et al.}\cite{perclos1998} is a valid metric for detecting drowsiness that has been used in many studies, e.g. Ji \textit{et al.} \cite{rti2002}, Flores \textit{et al.} \cite{iet2009}, Garcia \textit{et al.} \cite{garciaIV2012} and Dasgupta \textit{et al.} \cite{dasguptaITS2018}. Gaze detection based distraction warning is more complex than drowsiness detection. In the literature, two types of solutions were proposed: 1) geometric approaches and 2) machine learning approaches. The geometric approaches rely on the 3D gaze estimation via 3D modeling of face/eyes. As in the AttenD algorithm proposed by Ahlstrom \textit{et al.} \cite{gazeITS2013}, the estimated 3D gaze direction is compared with a predefined 3D safe region to detect distraction events. Vicente \textit{et al.} \cite{gazeTracking2015} compute the intersection of the driver's 3D gaze line and the car windshield plane. An EOR (eyes off the road) event would be triggered when the intersection point lies out of the safe region. Machine learning based methods directly predict a gaze zone from face and eyes image detections, avoiding 3D gaze direction estimation, which can be disrupted by scenario changes. Fig. \ref{fig::dms} (c) shows an example of 15 divided gaze zones for gaze classification. Fridman \textit{et al.} \cite{gazeIET2016} and Naqv \textit{et al.} \cite{ali2018} utilize respectively a random forest algorithm on facial landmarks vector, and directly a VGG neural network, to classify the gaze zones, i.e. which zone the driver is looking at. Yoon \textit{et al.} \cite{yoon2019} upgrade this method by using two NIR cameras and residual DNN to improve the accuracy and the robustness. More detailed reviews on gaze detection and DMS could be found in Dong \textit{et al.} \cite{dongITS2010} and Akinyelu \textit{et al.} \cite{akinyelu2020}.     

Aside from drowsiness and distraction detections, the NIR camera could be used to detect a driver's vital signs, e.g. pulse rate, respiratory rate, etc, e.g. in Wang \textit{et al.} \cite{algoPPG2017}, Magdalena \textit{et al.} \cite{ppg2018}, Wang \textit{et al.} \cite{wangTBE2020}, and Kurihara et al.\cite{tipHR2021}.


\subsubsection{LWIR cameras for night vision systems}

Thermal cameras are particularly suitable for detecting pedestrians and animals at night. Before the era of deep learning, object detection follows a traditional pipeline as: candidate region proposal, feature extraction and machine learning based classification. Popular object detection methods, such as Haar feature-based cascade AdaBoost classifier \cite{viola2001}, SVM classifier \cite{SVMTutorial} etc, are still popular in thermal imagery understanding due to less computational cost. Fang \textit{et al.} \cite{tvt2004} manually design features from hotspots in a thermal image to train a SVM classifier to recognize pedestrians. Xu \textit{et al.} \cite{xuITS2005} employ a SVM classifier and a mean-shift tracker for pedestrian detection and tracking. Forslund and Bjarkerfur \cite{autoliv2014} present a large animal thermal image dataset gathered over an 8-year period of driving in various locations. A cascade AdaBoost classifier is applied for animal detection and warning driver assistance systems. Savasturk \textit{et al.} from Daimler \cite{savasturk2015} find significant benefits in vehicle detection by combining RGB stereo images with monocular thermal images. In a recent work \cite{chenTIV2019}, Chen \textit{et al.} discover that by feeding CCF (Convolutional Channel Features) to a cascade AdaBoost classifier, LWIR camera can outperform stereo vision in pedestrian detection.

\begin{figure*}[t]
  \centering
  \subfigure[]{
    \includegraphics[width=0.5\textwidth]{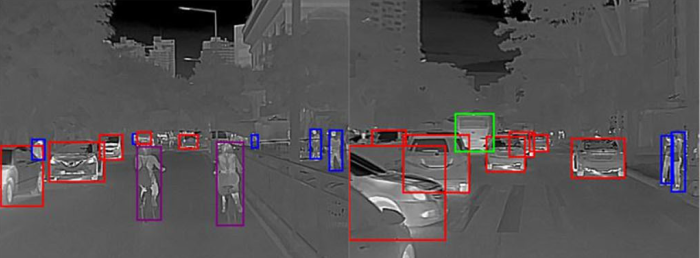}}
 \subfigure[]{
    \includegraphics[width=0.4\textwidth]{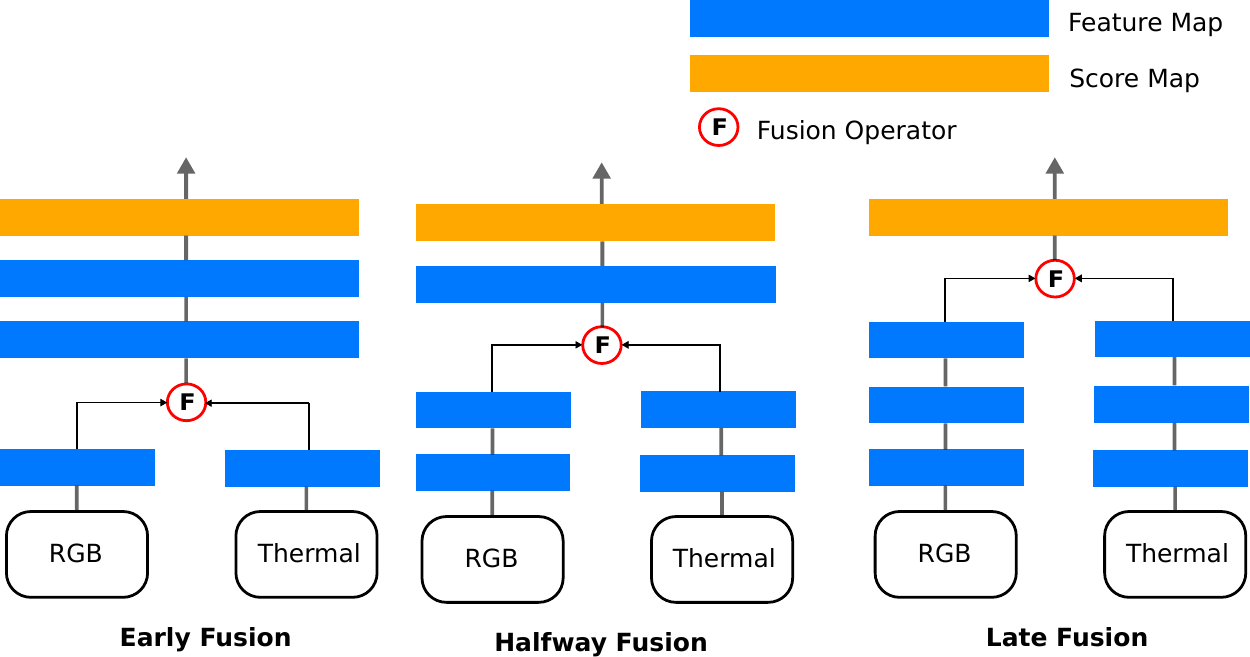}}  
\caption{(a) Object detection results from TIRNet. Pedestrian, cyclist, car, bus and truck are localized in 2D bounding boxes with different colors. (from \cite{dai2020}) (b) Different strategies of fusing RGB and thermal images according to specific stages.}
\label{fig::fusion}       
\end{figure*}

Entering into the era of deep learning when the CNNs (Convolutional Neural Networks) sweep all the computer vision benchmarks, there is no exception in thermal image processing. Kristo \textit{et al.} \cite{kristo2020} benchmark several popular object detectors, including Faster R-CNN \cite{fasterRCNN}, SSD \cite{ssd2016} and YOLOv3 \cite{yolov3}, that are retrained on a thermal image dataset for a surveillance system. YOLOv3 has been found to be significantly faster than other methods while still achieving comparable performance to the best. Dai \textit{et al.} \cite{dai2020} propose a TIRNet for pedestrian detection by modifying the SSD detector. The performance of TIRNet is reported better than YOLOv3 based on their annotated dataset and the KAIST dataset. Fig. \ref{fig::fusion} (a) shows several detection results of TIRNet. A large-scale thermal pedestrian dataset SCUT is presented in Xu \textit{et al.} \cite{scut2019}. Based on this dataset, the authors provide a detailed comparison between widely used detectors. Tumas \textit{et al.} \cite{tumas2020} present a ZUT dataset containing vehicle odometry and weather measures. Other than re-implementing the CNN detectors for thermal image processing, Grimming \textit{et al.} \cite{osa2021} dive deeply into a thermal camera's physical characteristics, and studied the relation between MTF (modulation transfer function), NETD, and the performance of fast R-CNN object detector \cite{fastRCNN}. 

In practice, fusing thermal and RGB images to deal with complex conditions is still indispensable for the majority of engineering systems \cite{fusion2019}. The fusion of DNNs from multispectral data could be done in different stages, as illustrated in Fig. \ref{fig::fusion} (b). The strategies can be roughly divided by the time to fuse: \textit{input fusion}, \textit{early fusion}, \textit{halfway fusion}, \textit{late fusion} and  \textit{score fusion}. Late fusion offers the flexibility
to directly fuse existing detectors inferring in parallel. Choi et al. \cite{icpr2016} and Park et al. \cite{pr2018} fuse this way two CNNs for proposal generation on color and thermal streams. With more modalities, Humblot-Renaux et al. \cite{renaux2020} investigate the late fusion for multispectral people detection from YOLO detectors, as well as Takumi et al. \cite{takumi17} from RGB, NIR, MIR and LWIR images. Other authors, e.g. Wagner \textit{et al.} \cite{esann2016}, Liu \textit{et al.} \cite{bmvc2016}, Li \textit{et al.} \cite{pr2019}, compare fusion schemes for pedestrian detection. The findings show that halfway fusion is superior to other approaches. As a result, the halfway fusion has become the default fusion strategy in CNN based multispectral image understanding, as demonstrated by Li \textit{et al.} \cite{bmvc2018}, Guan \textit{et al.} \cite{IFusion2019}, Zhang \textit{et al.} \cite{icip2020} and Yadav \textit{et al.} \cite{yadav2020}.

Another trend is looking for new neural network modules. Konig \textit{et al.} \cite{cvprw2017} use the faster-RCNN framework but with a new region proposal network (RPN) based on LWIR and RGB semantic segmentation. Zhang \textit{et al.} \cite{icip2020} propose a "Cyclic Fuse-and-Refine" module to optimize the complementary and consistency of multispectral features. In Li \textit{et al.} \cite{pr2019} and Guan \textit{et al.} \cite{IFusion2019}, illumination detection modules are proposed to dynamically assign the weights of multispectral features under a halfway fusion architecture. Zhang \textit{et al.} \cite{zliccv2019} propose an Aligned Region CNN (AR-CNN): a neural network module that compensates spatial misalignments of the features extracted from thermal-RGB image pairs. Dasgupta \textit{et al.} \cite{kinjal2021} extend the halfway fusion architecture with a multimodal feature embedding module (MuFEm) and a CRF-based Spatial-Contextual feature aggregation module.

Apart from object detection, thermal cameras could segment roads as well, e.g. in Pelaez \textit{et al.} \cite{pelaezIVS2015}, Yoon \textit{et al.} \cite{YoonIVS2016}, Humblot-Renaux \textit{et al.} \cite{renaux2020}.


\begin{figure*}[t]
  \centering
  \includegraphics[width=0.8\textwidth]{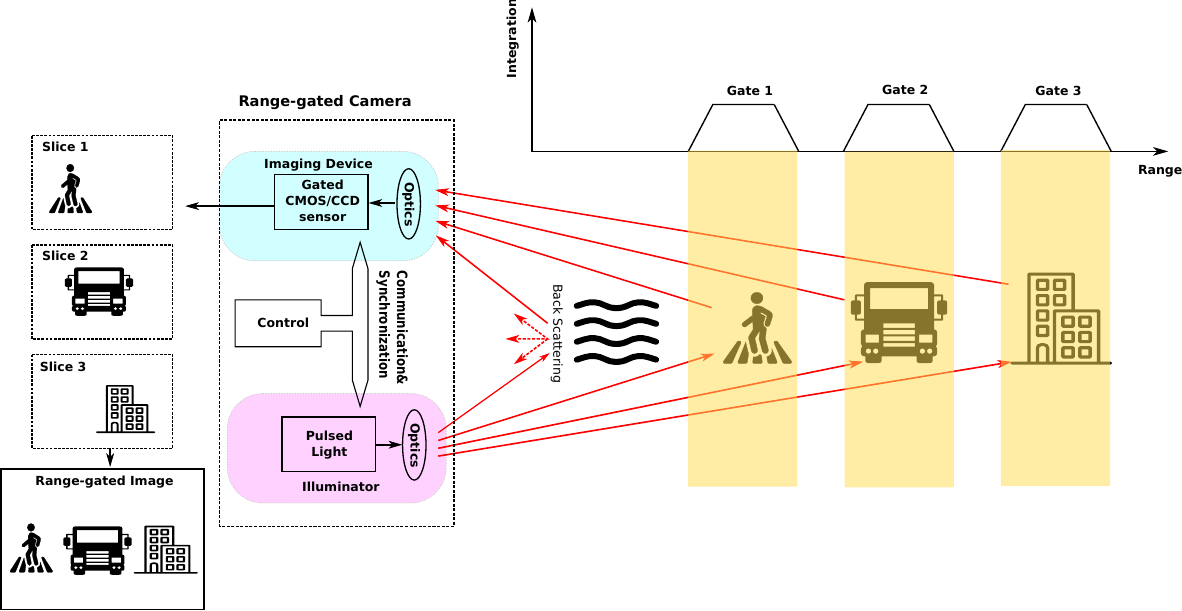}
\caption{The imaging principle of a range-gated camera: merging several image slices into a final image. A very short exposure defined by a gating function yields an image slice. The backscattering interference outside the range defined by a gating function has almost no effect on the outcome.}
\label{fig::gated_principle}       
\end{figure*}

\section{Range-Gated Camera}\label{sec::gated_camera}
To enhance the imaging quality under harsh conditions, range-gated imaging technique was firstly proposed in 1960s \cite{gated1966} and has been applied in night vision system \cite{gated2004}, submarine vision \cite{marin2020}. In recent years, range-gated cameras have gained popularity because of their resistance to adverse conditions \cite{marioCVPR2020}. 

\subsection{Principles}
A range-gated camera is an active imaging system in which an illuminator transmits pulsed light, and an image sensor is precisely synchronized to image the reflected lights within certain defined "gates". A general principle of a range-gated imaging system is shown in Fig. \ref{fig::gated_principle}. In the illuminator module, light pulses are emitted to illuminate the environment within the lens's field-of-view. Parts of the transmitted lights will be reflected by the surfaces of the objects and then be captured  partially by the receiving optics. Because the objects are at different ranges, the reflected photons are captured at different times. Unlike conventional cameras' exposure methods (global shutter or rolling shutter), a gated camera employs several \textit{gate functions} to expose the photons arriving at different times. Therefore, only the light arriving within the right timing window contributes to the final image. Usually, the exposure gates are very short: in the order of 0.01 - 2$\mu$s. As the example in Fig. \ref{fig::gated_principle}, three programmed gated functions generate three image slices containing objects at different ranges. The final image is obtained by merging those image slices. The components are introduced as follows:             

\textit{Illuminator} is triggered by the gating signals from a controller. Owing to narrow spectral width and high peak power, the laser is preferred over other kinds of lights. Different laser wavelengths ranging from visible, NIR to SWIR wavelengths could be applied. The NIR laser is popular because of its maturity and cost. For instance, 808nm laser is used in David \textit{et al.} \cite{David2006} and Spooren \textit{et al.} \cite{nick2016}. When considering better penetration in long-distance through fog or smoke, SWIR laser is preferred because it can achieve much higher transmission power while still meeting the eye safety standards. In \cite{gated2004}, a range-gated imaging system based on a Nd YAG laser at 1571nm reaches a 10km detection range. Similarly, in Baker \textit{et al.} \cite{OE2021}, a range-gated SWIR (1527nm) camera successfully penetrates heavy rains and detects obstacles 10km away. 

\textit{Gated image sensor}: The gated image sensors can perform multi-integration to generate a merged image by using gated signals. Due to the extremely short integration time, the gated image sensor has to be highly efficient. In Spooren \textit{et al.} \cite{nick2016}, a gated RGB-NIR image sensor with high NIR quantum efficiency (~40\%) is built. In Rutz \textit{et al.} \cite{apdMatrix2019}, a high-gain avalanche photodetector (APD) array containing $640\times 512$ InGaAs pixels is coupled with a SWIR laser transmitter. When operated in Geiger mode, the APDs become single-photon avalanche diodes (SPADs), meaning that even a single photon could trigger the avalanche effect. Burri \textit{et al.} \cite{spad2014} present a $512\times128$ pixel CMOS SPAD sensor capable of operating within an exposure window as small as 4ns. In Morimoto \textit{et al.} \cite{optica2020}, a 1M pixel CMOS SPAD image sensor is built for 3.8ns gating time.          


\subsection{Applications of in ADAS/AD}\label{ssec:gated-applications}
\begin{figure*}[h]
  \centering
  \includegraphics[width = 0.8\textwidth]{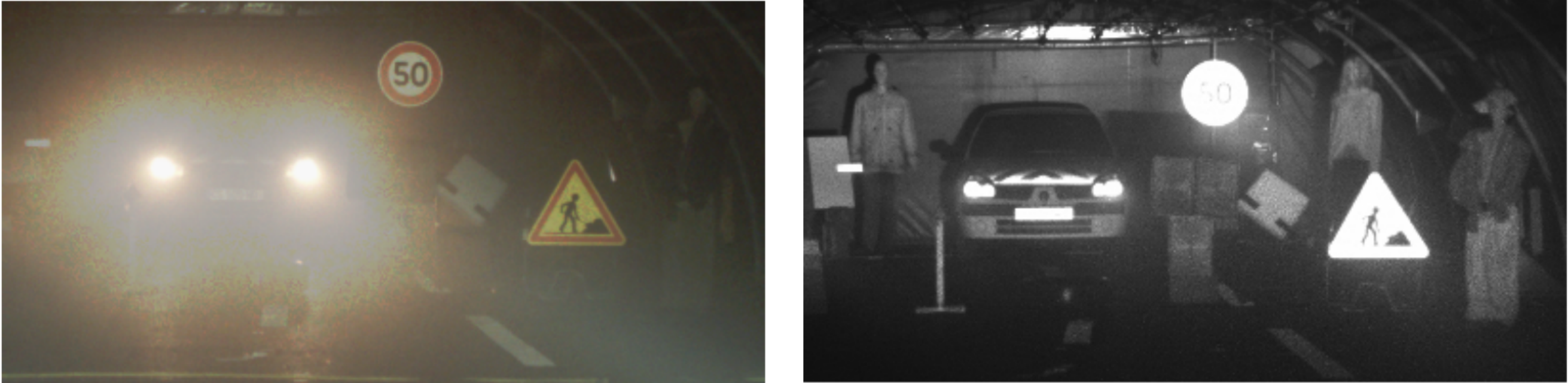}
\caption{Imaging results of a conventional RGB camera (left) and a range-gated camera (right) in an artificial fog (from \cite{marioIV2018}). Due to the backscattering effect caused by the fog, many objects are obscured in the RGB camera image, while the range-gated camera image is almost immuned to the fog. Meanwhile, the headlamp of the target car causes a strong blooming effect in the RGB camera image but has no impact on the range-gated camera image.}
\label{fig::gate_in_fog}
\end{figure*}

The range-gated cameras, as active sensors, are better suited to low-light conditions, such as a country road at night. Furthermore, owing to its imaging mechanism, only photons received at appropriate times are utilized for imaging. Such an attribute has two advantages: (1) No blooming effect when the photons from highly reflective objects do not fall within the sampling range. For example, the oncoming vehicles' headlamps have almost no impact on range-gated images. (2) Resistance to backscattering environments, such as fog/rain/smoke. A key parameter deciding the quality of a range-gated image is the modulation contrast:
\begin{equation}
  \centering
  \begin{split}
    Contrast &\simeq \frac{I_{target} - I_{background}}{I_{target} + I_{background} + 2I_{bsc}}
    \end{split}
  \label{eq::contrast}
\end{equation}
Where $I_{target}, I_{background}, I_{bsc}$ are the luminance of the target, background, and due to back scattering effect. Hence, the image quality in back scattering condition is defined by the strength of $I_{bsc}$, which can be calculated as:
\begin{equation}
  \centering
  I_{bsc} = \int_{2\gamma R_{on}}^{2\gamma R_{off}}\frac{PGe^{-X}\gamma^2}{2F_n^2\theta^2X^2}dX
  \label{eq::ibackscatter}
\end{equation}
where $R_{on}, R_{off}$ define a range interval during one exposure.  $G$ is the backscatter gain, $\gamma$ is the atmospheric attenuation coefficient, $\theta$ is the laser beam divergence, $P$ is the laser power, $F_n$ is the speed of the lens, and $X$ is the integration variable. Comparing with a conventional camera that all the backscattering photons are counted, a range-gated camera only perform the photon integration during a very short opening time, i.e. between $R_{on}$ and $R_{off}$, so that a higher contrast defined in Eq. \ref{eq::contrast} can be achieved. A comparison between a RGB camera and a range-gated camera in a fog environment is shown in Fig. \ref{fig::gate_in_fog}. Walz \textit{et al.} \cite{gateITSC2020} benchmark multi-model sensors in a well controlled artificial fog chamber. Both the quantitative and qualitative results show the superiority of range-gated cameras in such harsh conditions. Owing to the excellent performances in adverse conditions, the range-gated camera has the potential to be a strong competitor to infrared cameras, and has gained recognitions in recent years. On the industry side, \cite{LEDGated2010} first apply a NIR range-gated camera to aid driving at night. Grauer \textit{et al.} \cite{yoav2014} and \cite{yoav2015} present a high resolution (1.2M pixel) range-gated camera based on NIR VCSEL laser (808nm) and a gated CMOS image sensor. This sensor is suitable for use in active safety systems such as vulnerable object detection, forward collision warning, lane departure warning, traffic sign detection, etc.        

From 2017, a series of works around range-gated camera images are developed within the EU-founded DENSE project \footnote{\url{https://www.dense247.eu/}}. Supported by this project, the DENSE dataset \footnote{\url{https://www.uni-ulm.de/en/in/driveu/projects/dense-datasets/}} containing multi-model sensors (a range-gated camera, a RGB stereo camera, a LWIR camera, and a LiDAR) is released to the public. The dataset covers snow, rain, urban and sub-urban scenarios. The DENSE dataset is further annotated in Julca-Aguilar \textit{et al.} \cite{gated3D2021} as Gated3D dataset, in which more than 100K objects in 4 classes are manually annotated over 12997 image frames. Based on these datasets, Tobias \textit{et al.} \cite{gated2depth2019} present a deep neural networks (DNN) named as "\textit{gated2depth}", which can estimate th1e depth of each pixel in the range-gated camera. The proposed DNN architecture utilizes all the three slice images. Walz \textit{et al.} \cite{gateITSC2020} extend \textit{gated2depth} by incorporating aleatoric uncertainties into the pixel-wise depth estimation. Bijelic \textit{et al.} \cite{marioCVPR2020} propose a fusion neural network for adaptively fusing LiDAR, RGB camera, gated camera and radar features in an entropy estimation framework (higher entropy indicates more confidence). A delicate feature exchange network is designed to dynamically allocate the best features for each sensor. To explore the implied range information in the slice images, Julca-Aguilar \textit{et al.} \cite{gated3D2021} propose a DNN for 3D object detection. The proposed DNN is tailored to the temporal illumination cues from the three image slices. Based on the \textit{Gated3D} dataset, they demonstrated that using temporal cures from a range-gated camera, the 3D object detection results outperform a pure RGB based detection method.

\begin{figure*}[t]
  \centering
  \subfigure[]{
    \includegraphics[width=0.28\textwidth]{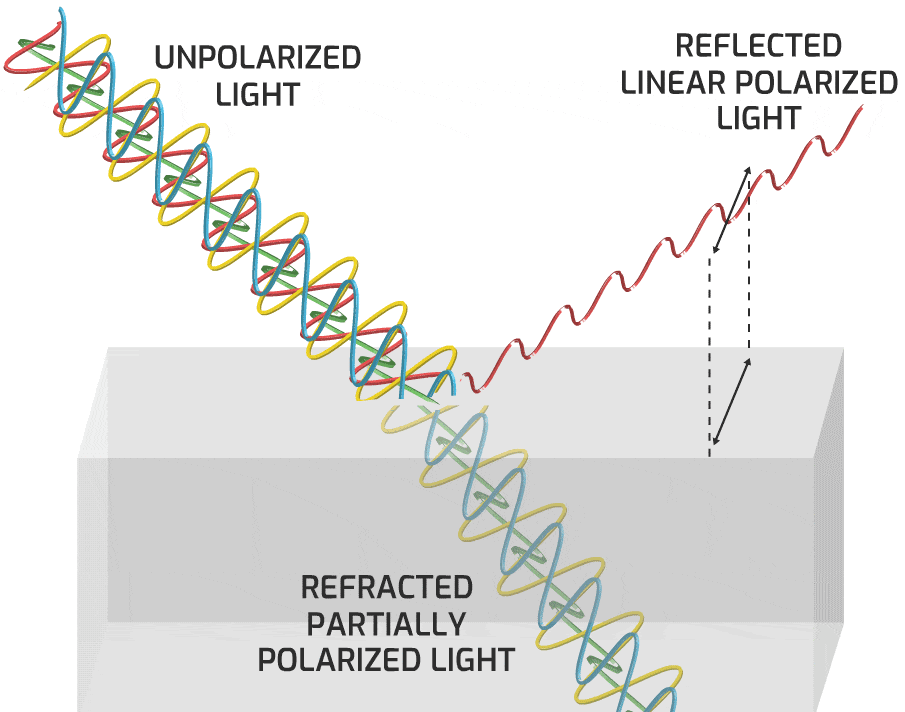}
  }
  \subfigure[]{
    \includegraphics[width = 0.28\textwidth]{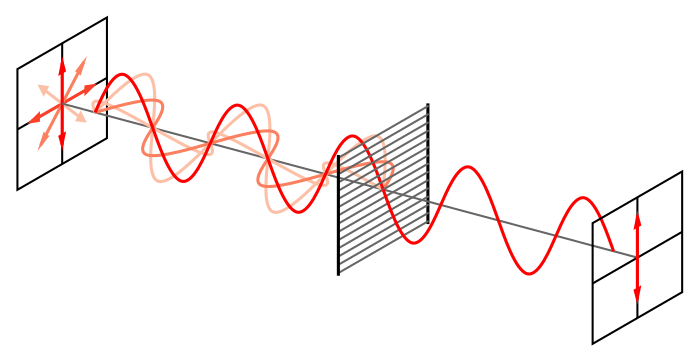}
  } 
    \subfigure[]{
    \includegraphics[width = 0.28\textwidth]{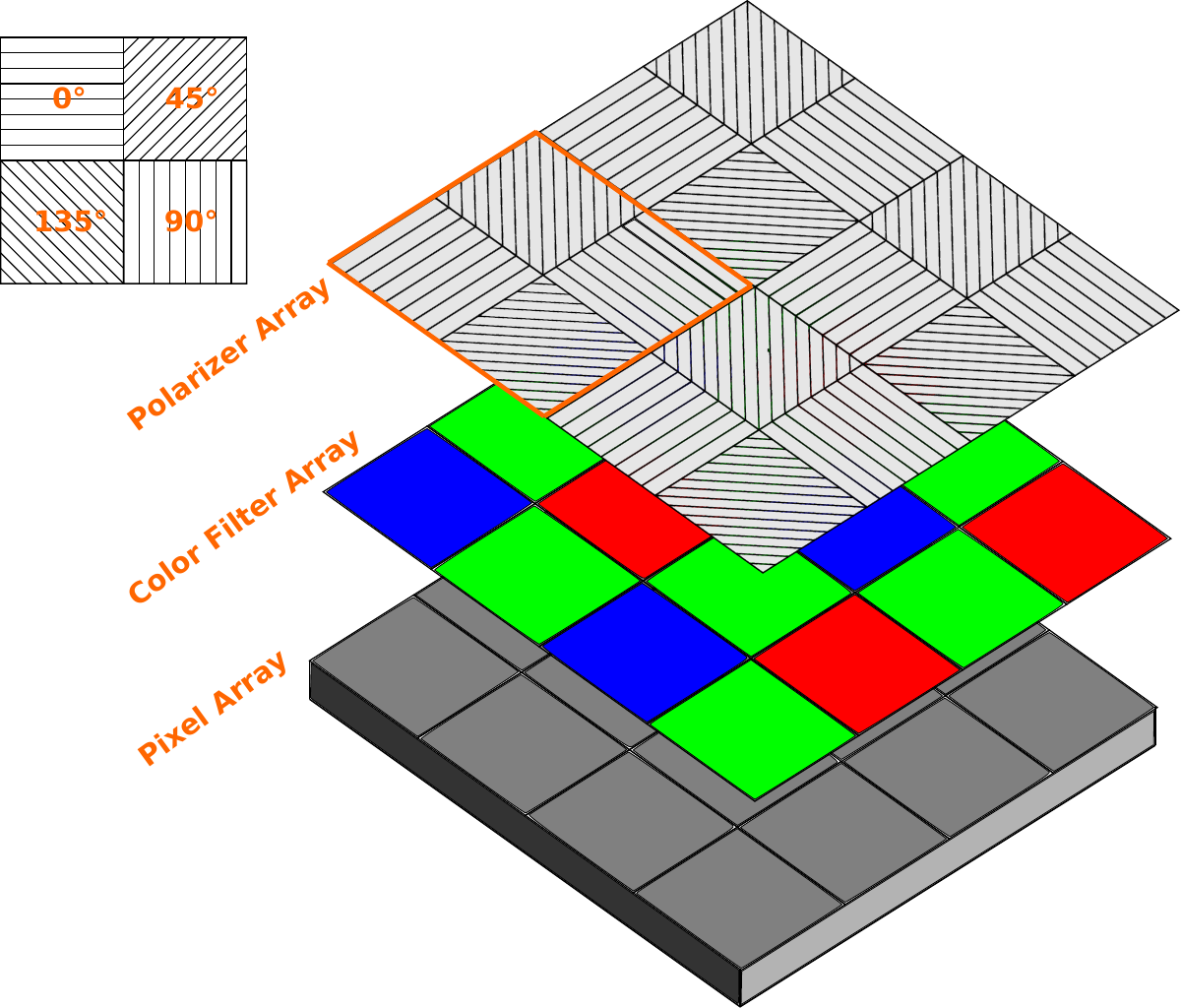}
  }
  \caption{(a) Unpolarized light beams could be converted to polarized light after reflection. (from \cite{lucid2018}). (b) A polarizer's function is to convert an unpolarized beam into a (linear) polarized beam (from \cite{wikiPolar}). (c) A schematic diagram of SONY IMX250 CMOS sensor. To acquire color and polarization information, a micro-polarizer array and a color filter array are placed on top of pixels.}
  \label{fig::polar}       
\end{figure*}
\section{Polarization camera}\label{sec::polar}
\subsection{Principle}

According to Sec. \ref{sec::light}, light passes through a medium as a \textit{transverse} wave, i.e. oscillating perpendicularly to the direction of propagation, that consists of an oscillating electric field and a magnetic field. For computer vision applications, only the electric field is considered. \textit{Polarization} is a fundamental and distinct property that describes the orientation of the light oscillation \cite{polarSN2018}. There are in general three kinds of polarized light: \textit{totally polarized} (\textit{linear, circular} or \textit{elliptic}), \textit{partially polarized} and \textit{unpolarized}. The majority of the light sources, e.g. the sun, streetlamps, emit unpolarized light, i.e. it vibrates randomly in all directions. 

Although most natural light is unpolarized, it can be converted to polarized light through the reflection from certain surfaces. In an ideal situation when the incident angle of unpolarized light is the angle of Brewster, according to Fresnel equations \cite{polarAP1990}, the reflected light is linear polarized (as shown in Fig. \ref{fig::polar} (a)). Otherwise, it would be partially polarized. Reflections from most flat surfaces are partially polarized as a function of incident angle. A more controllable way to obtain polarized light is to use a \textit{polarizer}, which is an optical filter that passes only specific polarized light while blocking light from other polarizations, as shown in Fig. \ref{fig::polar} (b). 

A concise representation of polarized light is the Stokes vector $\mathbf{S} $ \cite{hbOptics1996}, consisting of 4 parameters: $\mathbf{S} = [S_0, S_1, S_2, S_3]$. $S_0 (>0)$ is the total light intensity, $S_1$ and $S_2$ roughly represent the degree of linearly polarized light ($S_1$ stands for horizontal or vertical linear polarization, $S_2$ stands for 45° or 135° linear polarization). $S_3$ stands for ellipticity, which is usually ignored in applications.
\begin{equation}
  \begin{split}
    &S_0 = I_0+I_{90}=I_{45} + I_{135}\\
    &S_1 = I_0 - I_{90}, \quad S_2 = I_{45} - I_{135}
  \end{split}
  \label{eq::stokes}
\end{equation}

where $I_0, I_{45}, I_{90}$ and $I_{135}$ are the optical intensities at the corresponding  polarization direction, i.e. 0°, 45°, 90° and 135°. Other important physical properties, e.g. \textit{angle of polarization} (AoP) and the \textit{degree of polarization} (DoP) can be inferred from the Stokes vector as:
\begin{equation}
  AoP = \frac{1}{2}\times arctan(\frac{S_2}{S_1}), \quad DoP = \frac{\sqrt{S_1^2+S_2^2}}{S_0}
  \label{eq::polar_dopb}
\end{equation}
Varying between 0° and 180°, AoP represents the predominant axis of the light vibration. DoP is the ratio of the intensity of the polarized portion to the total intensity. For instance, a linearly polarized light has a DoP of 1, natural light usually has DoP between 0 to 0.5. 

Creating a practical and convenient polarimetric imaging system is not a easy work. In early research, Morel \textit{et al.} \cite{morel2005} make a polarization camera by manually rotating a polarizer in front of a normal camera. Three images are taken at different rotating angles of the polarizer to determine the Stokes vector for each pixel. In Wolff \textit{et al.} \cite{polarSplit1994}, a polarizing beam splitter is placed in front of 2 cameras so that the reflected and the transmitted beams are utilized to compute the polarization of each pixel. However, those methods either require a special environment for imaging, or are too expensive.

Powered by on-chip polarizer technology, modern image sensors can simultaneously acquire polarization and color information through a single shot. For instance, inside SONY's Pregius IMX250 CMOS sensor (as shown in Fig. \ref{fig::polar} (c)), a Polarization Filter Array (PFA) composed of four various angled micro-polarizers (0°, 45°, 90°, 135°) is placed on top of the CFA and photodiodes. The Stokes vector (as in Eq. \ref{eq::stokes}) and RGB vector for each pixel can be interpolated by using a special demosaicing process afterward. Such snapshot technology has a price advantage that it has been employed in many computer vision studies.     

\subsection{Applications in ADAS/AD}
Polarization cameras are not yet commercialized for automotive usages. Nevertheless, as the snapshot P-RGB image sensors (e.g. SONY IMX250, IMX253) become more popular, more researchers are beginning to investigate the potential benefits of light polarization. Current research focuses on \textit{image enhancement, object detection} and \textit{semantic segmentation}. 

As discussed in Sec. \ref{sec::issues}, the specular reflection, high contrast regions, and adverse weather would degrade the image quality. Although many intensity-based solutions (e.g. Li \textit{et al.} \cite{YouLIHDR}, and Wang \textit{et al.} \cite{mta2018}) could alleviate these issues, polarization cameras offer a new perspective. Wang \textit{et al.} \cite{specular2017} utilize a polarization camera to remove specular reflection because the DoP of the specular reflection part is much larger than the part of diffuse reflection, when an unpolarized light beam is reflected. Polarization cameras can also achieve high dynamic range (HDR) imaging to solve the over/under-saturation in high contrast conditions. As proposed by Wu \textit{et al.} \cite{polarHDR2020}, the 4 micro-polarizer patterns have similar effects as 4 different exposure times. Therefore, by using multiple polarization images at known pixel-specific exposure times, the irradiance maps can be estimated and hence construct an HDR image.

\begin{figure*}[t]
  \centering
  \subfigure[]{
  \includegraphics[width=0.58\textwidth]{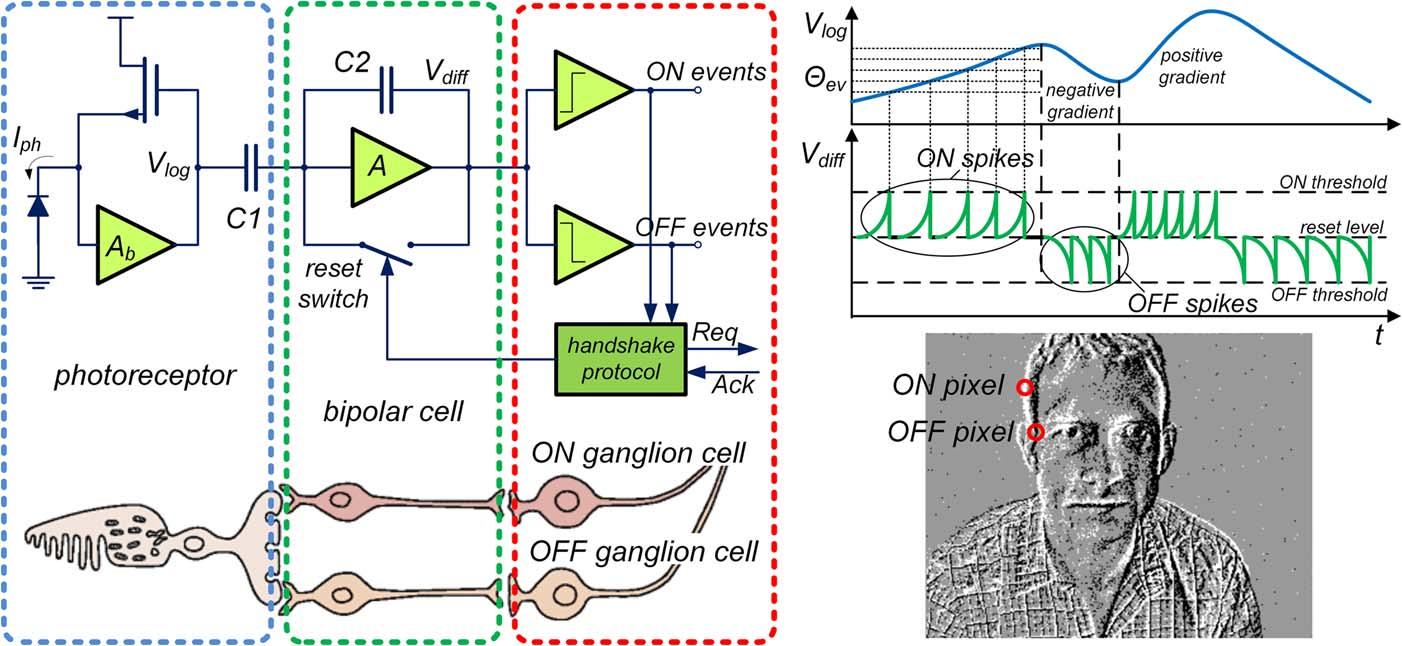}
}
\subfigure[]{
\includegraphics[width=0.38\textwidth, height = 5cm]{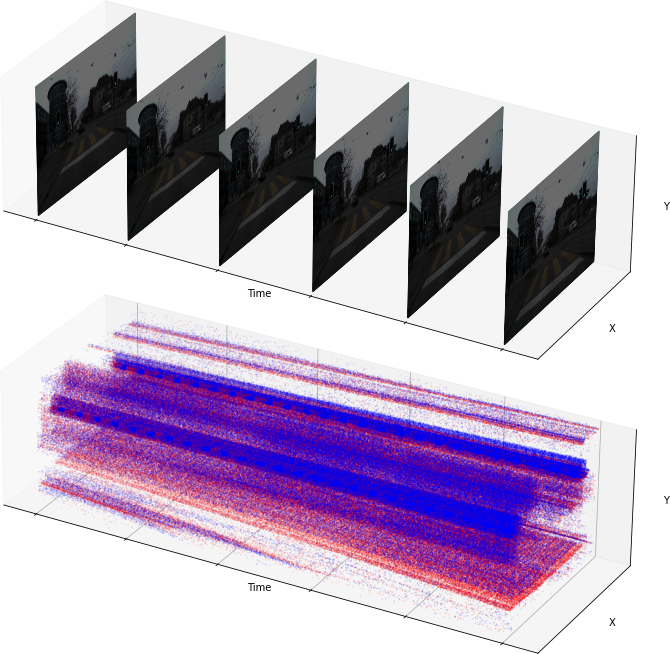}
}
  \caption{(a) Retinomorphic event vision with spiking output (from \cite{Posch14}). Human retina layers and corresponding dynamic vision sensor circuit for one pixel (left). Sample of signal waveform (top right), and response of an array of such pixels accumulated for a short time, for a sample scene showing a moving person (bottom right). (b) RGB frames from a standard camera (on top), compared to events flow (bottom), from the same time-synced scene issued from DSEC dataset~\cite{Gehrig21b}. Red points depict positive events, while blue points depict negative events.}
  \label{fig::event}       
\end{figure*}

Polarimetric images provide physical properties of the object, such as surface material and roughness, which can be utilized as a complement to traditional RGB image based object detection and segmentation. Wang \textit{et al.} \cite{polarCar2018} implement a feature selection process in polarimetric images and discover that the AoP is the most informative polarization feature. Then, for car detection, the AoP features are incorporated with a deformable part based models (DPM). The experimental results demonstrate that polarization features significantly reduce the false detection rate. Adding the polarization features to an object detection DNN, Blin \textit{et al.} \cite{blinITSC2019} and \cite{blinCVPRW2020} show that car detection results under adverse weather conditions could be improved by 20\% - 50\%. In addition, a new dataset PolarLITIS \cite{polarLITIS} containing RGB and polarimetric images under fog conditions was released to evaluate the performance gain of object detection from polarization information. The experiments in Blanchon \textit{et al.} \cite{visapp2019} and Xiang \textit{et al.} \cite{polarSeg2021} both find that the semantic segmentation for car and windows is largely improved thanks to the polarization features.

\section{Event camera}\label{sec::event}
In dynamic and unpredictable environments, traditional cameras would give blurry images or under/over-exposed images. The neuromorphic vision sensor is a good choice for a robust perception system. A general survey on event cameras is given in \cite{Gallego20}, and a tutorial aiming at some common processing methods applied for autonomous driving is given in \cite{Chen20b}. This section is complementary to these papers in providing a review of event vision for driving applications.

\subsection{Principles}
The ev0ent camera is also called \textit{address-event representation silicon retina}, \textit{neuromorphic}, or \textit{retinomorphic} camera, because it is inspired by eye retina, as described by \cite{Posch14}.
In the retina, the fundus of the eye, are located the cones and rods cells, that are sensitive to light, followed by layers of neurons.
Photosensitive cells convert light into electric signal transmitted to nerve cells.
Some signal exchanges occur from each photosensitive cell up to two bipolar ganglion cells: when activated, the first one represents \textsc{on} pulse whereas the second one represents \textsc{off} pulse.
In summary, \textsc{on} cell activates when a spatiotemporal brighter change in contrast occurs, \textsc{off} cell activates when a spatiotemporal darker contrast change occurs.
The brain is able to interpret these voltage spikes to give to us our sight sense.
This process leads to the following advantages: \textit{Independence from absolute light level:} It can be seen as an automatic gain control from the retina and allows vision capabilities for a very wide range of brightness. \textit{Lightweight data encoding for fast transmission}. Spikes are emitted continuously to the brain, avoiding the need to encode absolute intensities, and giving a high temporal resolution.


An event camera is designed to imitate the retina by bionomic pixel circuits (as shown in Fig.~\ref{fig::event} (a)), and hence inherit these advantages. Pixel outputs of an event camera are independent, they represent signed spikes as long as the photosensor observes a log-intensity difference above a threshold. The rate of following spikes of the same sign is an indication of the brightness change speed. Then, a stream of events is a sequence of timestamped signals, where each signal represents a positive or negative pulse (that is respectively, a state change to be more or less bright) for one or several points of the matrix sensor. An event camera does not stream full image frames in the way a conventional camera does at a given framerate. It acts in an asynchronous way with a very high temporal resolution and low latency, in an order of microseconds. The output difference between both sensors is shown in Fig.~\ref{fig::event} (b). Similarl to conventional image sensors, event sensors are made of Silicon and are sensitive to visible and NIR light. On the contrary, event cameras are often made without IR cut filter in order to gather more light. However, the use of specific wavelength filters may be necessary for certain applications. Modern neuromorphic cameras reach HD resolution, such as: Prophesee Gen4 CD ($1280\times720$ pixels) \cite{Finateu20}, Samsung DVS-Gen4 ($1280\times960$ pixels) \cite{Suh20}, CelePixel CeleX-V ($1280\times800$ pixels) \cite{Chen19a}. Some event cameras (e.g. iniVation DAVIS346 \footnote{\url{http://inivation.com/wp-content/uploads/2020/09/DAVIS346.pdf}}. CelePixel CeleX-V incorporates additional circuits in order to simultaneously output conventional images (monochrome in most cases) and sensed events. Such design gives the advantage of data fusion at exact superimposition, while at the expense of increasing noises caused by residual currents brought by those additional circuits. Rare event cameras are able to output both RGB events and frames, as iniVation DAVIS346B-Color\footnote{\url{http://inivation.github.io/inivation-docs/Hardware user guides/User_guide_-_DAVIS_USB3_development_kit.html}}, which includes a Bayer filter array to estimate RGB channels. Sample data of RGB event camera is available through the \textit{Color Event Camera Dataset} (CED) \cite{Scheerlinck19}.

\subsection{Advantages}
\label{ssec:ev-advantages}
Event cameras are bio-inspired passive sensors that try to imitate millions years evolution of sight sense. General advantages of event camera are stated by \cite{Gallego20}:
\textit{Microsecond temporal resolution} for detection and timestamp. A direct consequence is the ability to always avoid motion blur as it exists for conventional cameras. Furthermore, the event camera outputs at sub-millisecond latency, which is approximately equivalent to a virtual $>1000FPS$ frame-based camera. \textit{Low power consumption}, in the order of $10mW$ to $100mW$ for typical event cameras, while usually between $1W$ and $3.5W$ for industrial RGB cameras. \textit{Broad dynamic range}: an event camera's dynamic range can easily reach $>120dB$ without a special design. In contrast, a normal RGB camera needs a dedicated pixel design to boost its dynamic range from typical $60-70dB$ to $110dB$.

\begin{figure*}[htbp]
  \centering
  \subfigure[There is a bike on right side.]{
    \includegraphics[width=0.4\columnwidth]{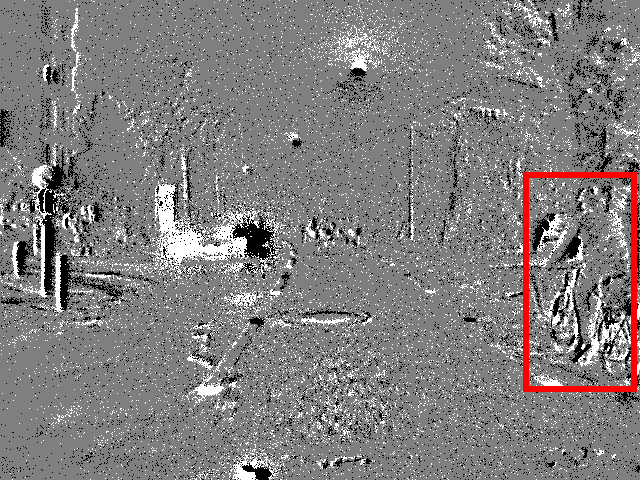}
    \includegraphics[width=0.4\columnwidth]{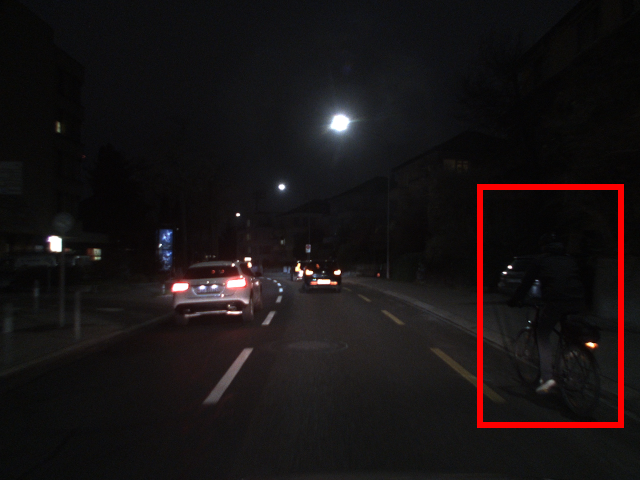}
  }
  \subfigure[What comes after the tunnel?]{
    \includegraphics[width=0.4\columnwidth]{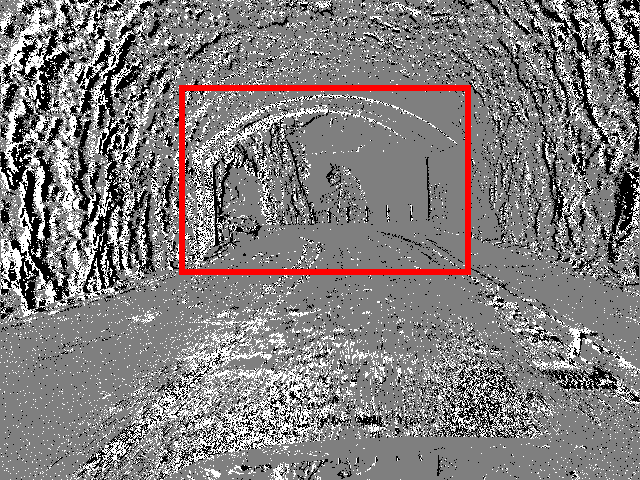}
    \includegraphics[width=0.4\columnwidth]{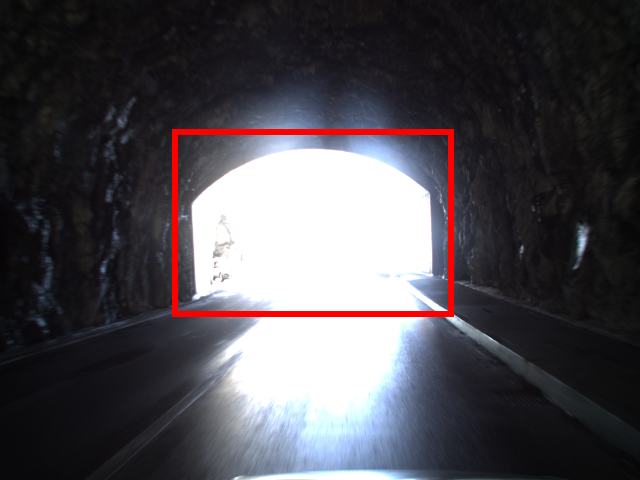}
  }
   \subfigure[Front car's speed equals ours so it is not visible in events.]{
    \includegraphics[width=0.4\textwidth, height = 2.7cm]{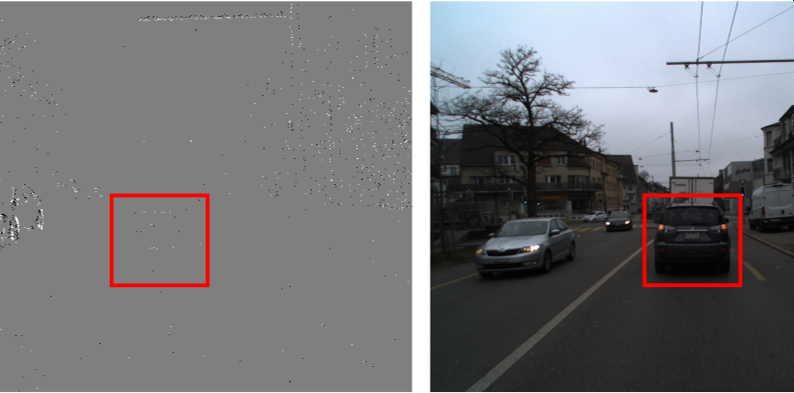}
    \label{fig::event_challenges_equmotion}
  }
  \subfigure[With events, shadows can look like noise or even obstacles.]{
    \includegraphics[width=0.4\textwidth, height = 2.7cm]{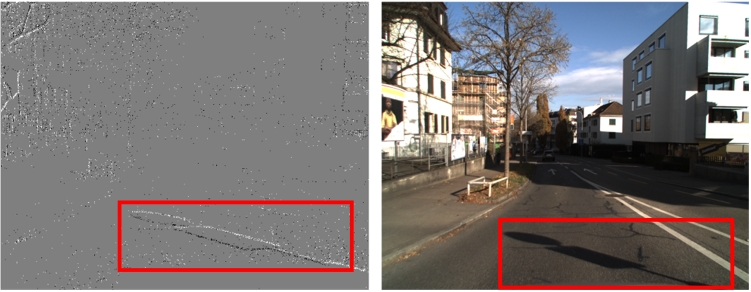}
    \label{fig::event_challenges_shadow}
  }
  \caption{Advantages and limits of event camera images compared to normal images: (a) Night scene. (b) Glare light. (c) Stationary car (d) Shadow on the ground. Scenes taken from DSEC dataset \cite{Gehrig21b}. }
  \label{fig::event_advantages}
\end{figure*}

All these advantages are desirable for intelligent vehicles: very high temporal resolution allows for detection of fast-moving elements of the scene; very low-latency is important for safety-critical applications; very high dynamic range allows to perceive in challenging lighting conditions. Event camera capabilities in driving scenes are illustrated in Figure~\ref{fig::event_advantages}.

\subsection{Data representation and processing}
\label{ssec:ev-processing}
Unlike frame cameras, neuromorphic cameras output stream of events $\mathcal{E} = \{ e \forall (x,y,t,p) \}$ that each event $e(x,y,t,p)$ encodes position $(x,y) \in \mathbb{N}^2$, timestamp $t$, polarity of the brightness change $p \in \{-1,+1\}$. In signal processing point of view, an event $e(x,y,t,p)$ can be considered as a continuous function using diracs $e = p \cdot \delta(\xi - x, \upsilon - y) \cdot \delta(\tau-t)$ where $(\xi,\upsilon) \in {\mathbb{R}^+}^2$ represents 2D spatial positioning in pixel array and $\tau \in \mathbb{R}^+$ represents continuous running time. There are two ways of processing event flows: asynchronous processing when an event arise, that is event-by-event processing, or accumulation of events within a temporal window, i.e. process them as an array or as a tensor.

\textit{Event-by-event processing:} it is a natural way to keep the raw asynchronous and sparse event(spike) flow, whereas current computers are not designed for spikes processing. Standard processors architectures (CPUs and GPUs) are good to process dense arrays of data, but are not able to process irregular flows of independent events at a very high rate. Specific biologically inspired hardware is designed to efficiently process event-by-event, such as \textit{ROLLS} processor \cite{Qiao15}, IBM \textit{TrueNorth} chip \cite{Akopyan15}, and Intel \textit{Loihi} chip \cite{Davies21}. These spike processors are particularly interesting since they open the door for hardware SNN (spiking neural networks) with low power consumption. SNNs are designed to imitate brain neurons and are the most popular and direct way to process event-by-event flows. With or without specific hardware, some early investigations about SNNs have been done. However, as a single event gives insufficient information for understanding, new events are used iteratively to update a system's state. While some methods apply standard optimization techniques or filters (such as \cite{Gehrig20b, Akolkar20, Nunes21}), most of them integrate asynchronous events in artificial neural networks. SNNs have already been proposed for many applications. For example, \cite{Osswald17, Barbier21} for stereo depth estimation, \cite{Viale21} for classification, \cite{Paredes20} for optical flow, \cite{Parameshwara21a} for background motion separation, \cite{Kreiser18a, Kreiser20} for heading estimation and loop closure detection, \cite{Stagsted20a, Vitale21} for robotic control, \cite{Ruckauer19} for target following, \cite{Salvatore20} for collision avoidance (for drone).


\textit{Events binning:} It is generally more practical to process batches of incoming events rather than processing individual events. Usually, successive events are gathered and compressed into a dense array or a tensor that is similar to an image frame. Both arrays and tensors can be efficiently processed by standard computer hardware. Binning events cause additional latency. However, this drawback is moderate and acceptable as the general advantages are still kept. There are two strategies of events binning: \textit{via a time window}, or \textit{via a queue of a fixed number of events.} The usage of time windows is easy and common, while can lead to accumulated arrays without or too many events. Hence, the sampling time should be tuned accordingly and gives a synchronous process. Zou \textit{et al.}\cite{Zou17} use adaptive accumulation time (making the method asynchronous), Rebecq \textit{et al.}\cite{Rebecq17b} use overlapping pairs of time windows, and Joubert \textit{et al.}\cite{Joubert19} combine time windows of different lengths. The study \cite{Akolkar15} contains a performance analysis of various time window durations for a classification task. The strategy of binning by a fixed number of events is used in \cite{Moeys16, Moeys18, Afshar19, Paredes21}. It allows for keeping accumulated representations with similar appearances (same density of events) and for asynchronous process. Nevertheless, the following operations should be fast enough when huge quantities of events arrive in a short time.

After accumulating enough events, the next issue is \textit{encoding}, i.e. extracting effective event attributes. Several \textit{hand-crafted encoding} methods have been proposed in the literature. For example, leveraging "frequency encoding" representation from Chen \textit{et al.}\cite{Chen18}, where a standard YOLOv3 CNN architecture \cite{yolov3} is utilized for pedestrian detection. Chen \textit{et al.}\cite{Chen20a} also get the best results with the "frequency encoding" among other encoding schemes for driver monitoring applications. Perot \textit{et al.} \cite{Perot20} test different accumulation and encoding strategies for object detection, with the best results using the "discretized event volume" representation from Zhu \textit{et al.}\cite{Zhu19a}. Besides human-designed features, generalized expression can also be learned automatically in an \textit{end-to-end} manner. Tulyakov \textit{et al.}\cite{Tulyakov19} model events as a stream of sparse 3d data points, and then apply a MLP (Multi-Layer Perceptrons) to learn an optimal encoding for a stereo-matching problem. Experimental results show that the learning-based encoding is better than the best hand-crafted approach. Cannici \textit{et al.}\cite{Cannici20} propose specific LSTM (Long Short-Term Memory) recurrent modules as a flexible way to learn task-dependent event-surfaces, and show better performance in optical flow estimation. Li \textit{et al.}\cite{Li19} apply a SNN to encode events and generate visual attention maps for further fusion with frame images, in an object detection framework.

At last, event accumulation can be motion compensated with a fast algorithm, typically using a joint IMU inside event camera \cite{Rebecq17b}. This guarantees for event accumulation with no blur effect in case of a long accumulation time.

\subsection{Applications in Autonomous Driving or ADAS}
The event camera is especially useful for systems running with real-time interactions, non-controlled enlightenment conditions, and low latency. In this paper, we focus on their application in the autonomous driving field.

\subsubsection{Dataset in driving scenes}
To apply the event cameras in autonomous vehicles, large and well-annotated datasets are indispensable. The neuromorphic vision community is very active in it. Because event cameras are still in early stages, many published datasets (e.g. MVSEC \cite{Zhu18a}, DDD17 \cite{Binas17}, etc.) in recent years are still in low image resolution (e.g. less than 640$\times$480) due to hardware limits. The first HD event camera released in public is the CelePixel CeleX-V in 2019 \cite{Chen19a}. Larger resolution benefits further object detection range and better recognition for small objects, while  poses challenges for computation capability because of huge event flows. We expect to see more and more HD event camera datasets as Perot \textit{et al.}\cite{Perot20} appear in public. The published datasets are for various purposes, such as target detection \cite{Binas17}, lane detection \cite{Cheng19}, drowsiness detection \cite{Chen20a} etc. A comprehensive summary of the current open datasets is demonstrated in Table~\ref{tab::dataset_event}.

\subsubsection{Object detection}
Object detection is a traditional but critical topic for autonomous driving systems. Since the event camera is a new sensor, labeled event datasets are scarce. Leveraged by the pseudo-labels warped from frame images in DDD17 dataset, Chen \textit{et al.}\cite{Chen18} apply several popular CNN object detectors in event camera images, and achieved good results in motion-blurring scenarios. The PKU-DDD17-CAR dataset, i.e. the annotation of DDD17 by Li \textit{et al.}\cite{Li19}, is used by Cao \textit{et al.}\cite{Cao21} to detect vehicles. Hu \textit{et al.}\cite{Hu20b, Hu21} augment respectively parts of a day and a night sequence of MVSEC dataset \cite{Zhu18a} with car annotations for DNN training. Except for events-only object detection, another direction is to fuse frame images for better performance. Li \textit{et al.}\cite{Li19} apply a SNN for the event stream to generate attention maps that feed to a CNN concatenated to standard frames, as in an early fusion scheme. Cao \textit{et al.}\cite{Cao21} fuse events and frames at different encoding levels from parallel heads using feature attention gate components. Hu \textit{et al.}\cite{Hu20b, Hu21} illustrate proposed grafted networks and events synthesis from video frames with a car detection use case. Pedestrian detection is also important and is explored in \cite{Chen19b, Jiang19, Cladera20, Wan21}, in which individual datasets are utilized according to specific cameras. Chen \textit{et al.}\cite{Chen19b} compare different accumulation methods coupled with early fusion and late fusion schemes. In Jiang \textit{et al.}\cite{Jiang19}, events and frames data channels are both fed into different CNNs and then fuse multiple confidence maps to achieve good pedestrian detection. Cladera \textit{et al.}\cite{Cladera20} implement a BNN (binary neural networks) on FPGA for fast detection. Wan \textit{et al.}\cite{Wan21} propose a Pedestrian-SARI dataset and alternative events representations for asynchronous CNN detection. Lane extraction problem is investigated in Cheng \textit{et al.}\cite{Cheng19}, in which a DET dataset, labeled lane markings in HD event camera, is released to be public. Meanwhile, several popular CNN-based lane extraction algorithms are benchmarked and the results show good performances. A more general object detection method and an annotated dataset are proposed in Perot \textit{et al.}\cite{Perot20}, where a CNN combined with a LSTM is used to keep detections when movements stop. The proposed method is evaluated for HD events road scenes, released in their 1 Megapixel Automotive Detection Dataset.

\subsubsection{Motion segmentation} motion segmentation or moving object detection by an event camera is more convenient than a conventional camera. This topic is addressed in \cite{Mitrokhin18, Stoffregen19a, Zhou21a, Parameshwara21b}. In general, those approaches compensate first camera motion as background movement, as it is likely to cause the most prominent number of events. Then, moving objects are segmented through different clustering strategies. For instance, Mitrokhin \textit{et al.}\cite{Mitrokhin18} group events into clusters via morphological operators, then tracks the multiple moving objects. Stoffregen \textit{et al.}\cite{Stoffregen19a} warp the events several times to cluster moving objects. Zhou \textit{et al.}\cite{Zhou21a} cluster the objects via graph cut on linked space-time event graph. \cite{Parameshwara21b} cluster the objects with split and merge strategy and track grouped events. Monda \textit{et al.} \cite{Mondal21} don't consider background motion. Instead, moving objects are segmented from a fixed event camera flow, and then grouped by a k-NN graph method.

\subsubsection{Driver monitoring system} Currently, few investigations have been done with event cameras to monitor driver status. Chen \textit{et al.}\cite{Chen20c} focus on drowsiness detection, and compare some classification algorithms on their event dataset. Provided with a new dataset, Chen \textit{et al.}\cite{Chen20a} compare different CNN architectures and events accumulation schemes for driver drowsiness detection, gaze-zone recognition, and hand-gesture recognition.


\begin{table*}[t]
  \centering
  \caption{Several typical open datasets for multiple sensing modalities (infrared, gated infrared and polarization cameras)}
\begin{tabular}{l c c c c c c cccc}
    \toprule
  \multirow{2}{*}{Name} & \multicolumn{5}{c}{Modality} & \multirow{2}{*}{Size} & \multirow{2}{*}{Annotation} & \multirow{2}{*}{Location} & \multirow{2}{*}{Year} \\\cmidrule(lr){2-6}
   &RGB &NIR &LWIR &Gated &Polar &&&& \\\hline
  FLIR-ADAS\cite{FLIRADAS} &\checkmark& &\checkmark & &   &13K & \tabincell{c}{Person, car, bicycle\\dog, other vehicles} & US & 2020\\\hline
  KAIST\cite{kaistDS2018} &\checkmark & &\checkmark & & &  95K &\tabincell{c}{Person, Pedestrian\\ and cyclist}&Korea& 2015 \\\hline
  SCUT\cite{scut2019}  &  & &\checkmark & & &    211K & \tabincell{c}{Walk/ride/squat people.} & China &2019\\\hline
  ZUT\cite{tumas2020}   &  & &\checkmark & & &    110K& \tabincell{c}{9 classes including\\pedestrian, cyclist, animal} & EU &2020 \\\hline
  RANUS\cite{ranus2018} &\checkmark  &\checkmark & & &    &4K  &\tabincell{c}{10 classes including\\vehicle, road, pedestrian, vegetation} & Korea& 2018\\\hline
  SparsePPG \cite{ppg2018} &\checkmark  &\checkmark & & & &19 Video sequencs  &\tabincell{c}{With ground truth\\ driver PPG waveform} & US& 2018\\\hline
  DENSE \cite{denseDataset} &\checkmark & &\checkmark &\checkmark &  &13K &\tabincell{c}{4 classes including\\ pedestrian and car} &EU &2020\\\hline
  PolarLITIS \cite{polarLITIS} &\checkmark & & & &\checkmark &2.5K &\tabincell{c}{Car, Person, \\bike, motorbike} &EU & 2021\\\hline
  ZJU-RGB-P \cite{polarSeg2021} &\checkmark & & & &\checkmark &394 &\tabincell{c}{Pixel-wise semantic segmentation\\ for building, glass, car, pedestrian, road, etc} &China & 2021\\
  \bottomrule
\end{tabular}
\label{tab::dataset_thermal}
\end{table*}

\begin{table*}[htbp]
  \centering
\caption{Event camera datasets with driving scenes. Top part lists low resolution datasets ($<1280\times720$px), bottom part lists high resolution datasets ($\ge 1280\times720$px).}
\begin{threeparttable}
\begin{tabular}{l|ccccccc}
    \toprule
  Name & Pixel resolution & Other modalities & Aimed problems & Size & Annotations & Location & Year \\\hline
  \tabincell{l}{PRED18\cite{Moeys18} \textit{(includes}\\ \textit{previous PRED16\cite{Moeys16})}} & $240 \times 180$ & Grey frames\ssymbol{1} & \tabincell{c}{Mobile target\\following} & 1.25h & \tabincell{c}{prey size,\\\textit{prey position}} & \tabincell{c}{Northern\\Ireland} & \tabincell{c}{2018,\\\textit{2016}}\\ 
  \cline{2-8}
  \tabincell{l}{DDD20\cite{Hu20a} \textit{(includes}\\\textit{previous DDD17\cite{Binas17})}} & $346 \times 260$ & \tabincell{c}{Grey frames\ssymbol{1}\\IMU\ssymbol{1}\\Car data\\GNSS} & Vehicle control & \tabincell{c}{39h\\\textit{+ 12h}} & - & \tabincell{c}{USA,\\\textit{Swiss,}\\\textit{Germany}} & \tabincell{c}{2020,\\\textit{2017}} \\
  \cline{4-4}\cline{6-6}\cline{8-8}
  PKU-DDD17-CAR\ssymbol{2}\cite{Li19} & & & Detection & & "Car" & & 2019 \\
  \cline{4-6}\cline{8-8}
  Ev-Seg\ssymbol{2}\cite{Alonso19} & & & Segmentation & 20 intervals & Semantic seg & & 2019 \\
  \cline{2-8}
  N-Cars\cite{Sironi18} & $304 \times 240$ & - & Classification & 24K samples & \tabincell{c}{"Car"\\"Background"} & unknown & 2018 \\
  \cline{2-8}
  MVSEC\cite{Zhu18a} & \tabincell{c}{$346 \times 260$\\2 cameras} & \tabincell{c}{Grey frames\ssymbol{1}\\Grey stereo camera\\IMU\\LiDAR\\GPS\\Motion capture} & \tabincell{c}{Depth\\Localisation} & 1h & Depth & USA & 2018 \\
  \cline{4-4}\cline{6-6}\cline{8-8}
  MVSEC-OF\ssymbol{3}\cite{Zhu18b} &  &  & Optical flow &  & Optical flow &  & 2018 \\
  \cline{4-6}\cline{8-8}
  MVSEC-DAY20\ssymbol{3}\cite{Hu20b} &  &  & Detection & \tabincell{c}{partial seq.\\"\textit{outdoor\_day2}"} & "Car" &  & 2020 \\
  \cline{4-6}\cline{8-8}
  MVSEC-NIGHTL21\ssymbol{3}\cite{Hu21} &  &  & Detection & \tabincell{c}{partial seq.\\"\textit{outdoor\_night1}"} & "Car" &  & 2021 \\
  \cline{2-8}
  Slasher dataset\cite{Hu19} & $346 \times 260$ & \tabincell{c}{Grey frames\ssymbol{1}\\Steering\\Radio localisation} & Vehicle control & 2 sequences & - & Swiss & 2019 \\
  \cline{2-8}
  \tabincell{l}{Event Camera\\Driving Sequences\cite{Rebecq19}} & $640 \times 480$ & RGB camera & \tabincell{c}{Frames\\reconstruction} & 40 sequences & - & Swiss & 2019 \\
  \cline{2-8}
  CED\cite{Scheerlinck19} & \tabincell{c}{$346 \times 260$\\RGB event cam} & RGB frames\ssymbol{1} & \tabincell{c}{Color frames\\reconstruction} & 50min & - & unknown & 2019 \\ 
  \cline{2-8}
  \tabincell{l}{Pedestrian Detection\\Dataset\cite{Miao19}} & \tabincell{c}{$346 \times 260$\\RGB event cam} & - & Detection & 12 recordings & "Pedestrian" & China & 2019\\
  \cline{2-8}
  EDDD\ssymbol{4}\cite{Chen20c} & $346 \times 260$ & - & Driver monitoring & 260 sequences & Drowsiness & China & 2020 \\
  \cline{2-8}
  NeuroIV\ssymbol{4}\cite{Chen20a} & \tabincell{c}{$346 \times 260$\\RGB event cam} & \tabincell{c}{RGB frames\ssymbol{1}\\Depth maps\\NIR frames} & Driver monitoring & 27K samples & \tabincell{c}{Drowsiness\\Gaze-zones\\Hand-gestures} & China & 2020 \\
  \cline{2-8}
  GAD Dataset\cite{Tournemire20} & $304 \times 240$ & - & Detection & 39h & \tabincell{c}{"Car"\\"Pedestrian"} & France & 2020 \\ 
  \cline{2-8}
  \tabincell{l}{Brisbane Event\\VPR\cite{Fischer20}} & \tabincell{c}{$346 \times 260$\\RGB event cam} & \tabincell{c}{RGB frames\ssymbol{1}\\RGB camera\\IMU\ssymbol{1}\\GPS} & \tabincell{c}{Visual place\\recognition} & 8km & Landmarks & Australia & 2020 \\ 
  \cline{2-8}
  DENSE\ssymbol{5}\textsuperscript{,}\ssymbol{6}\cite{Hidalgo20} & $346 \times 260$ & \tabincell{c}{RGB frames\\Depth maps} & \tabincell{c}{Depth\\Segmentation} & 8K samples & \tabincell{c}{Depth\\Semantic seg} & - & 2020 \\ 
  \cline{2-8}
  DSEC\cite{Gehrig21b} & \tabincell{c}{$640 \times 480$\\2 cameras} & \tabincell{c}{2x RGB cameras\\LiDAR\ssymbol{4}\\RTK GPS\ssymbol{4}} & \tabincell{c}{Depth\\Localisation} & 53min & Depth & Swiss & 2021 \\
  \cline{4-4}\cline{6-6}\cline{8-8}
  DSEC-OF\ssymbol{7}\cite{Gehrig21c} & & & Optical flow & & Optical flow & & 2021 \\
  \cline{2-8}
  EventScape\ssymbol{5}\cite{Gehrig21a} & $512 \times 256$ & \tabincell{c}{RGB frames\\Depth maps\\Car data} & \tabincell{c}{Depth\\Segmentation} & 2h & \tabincell{c}{Depth\\Semantic seg} & - & 2021 \\ 
  \cline{2-8}
  Pedestrian-SARI\ssymbol{4}\cite{Wan21} & $346 \times 260$ & Grey frames\ssymbol{1} & Detection & 141 sequences & "Person" & China & 2021\\
  \hline\hline
  DET\cite{Cheng19} & $1280 \times 800$ & - & Lane extraction & 5h & Road lanes & China & 2019 \\
  \cline{2-8}
  1Mp Detection\cite{Perot20} & $1280 \times 720$ & - & Detection & 14h & \tabincell{c}{"Car"\\"Pedestrian"\\"Two-wheeler"} & France & 2020 \\
   \bottomrule
\end{tabular}
\begin{tablenotes}
\item[\ssymbol{1}] Available from the event camera itself.
\item[\ssymbol{2}] Extension of DDD17, providing ground truth to other problem.
\item[\ssymbol{3}] Extension of MVSEC, providing ground truth to other problem.
\item[\ssymbol{4}] Not available for download, might be available upon request to the authors.
\item[\ssymbol{5}] Simulated data.
\item[\ssymbol{6}] Distinct from DENSE dataset for LWIR and range-gated cameras \cite{denseDataset} presented in Table~\ref{tab::dataset_thermal} and in Section~\ref{ssec:gated-applications}.
\item[\ssymbol{7}] Extension of DSEC, providing ground truth to other problem.
\end{tablenotes}
\end{threeparttable}
\label{tab::dataset_event}
\end{table*}

\subsection{Remaining Challenges}
\label{ssec:ev-challenges}
Although the attributes of event cameras are attractive, they are still quite young, and hence suffer several restrictions for wide applications. The first restriction involves the optimal performance in complex enlightenment scenarios, for example, the camera biases and noise. Biases need to be carefully tuned to achieve optimal perception according to the conditions (scene brightness and dynamics, ambient temperature, admissible noise, etc). Tuning the event camera's parameters is not a straightforward task, as there are too many correlated parameters to adjust. Usually, some general tries are required at first to correct the parameters. Details on how to control an event camera are given in Delbruck \textit{et al.}\cite{Delbruck21}. Other constraints concern the sensing characteristic. The most typical issue is the relative static object, as illustrated in Fig. \ref{fig::event_advantages} (c), because the front car and ego-vehicle are both stopped, the event camera barely perceives the front car. Fortunately, such a problem could be overcome by applying RNN in object detection Perot \textit{et al.}\cite{Perot20}. Irregular data bandwidth is another constraint caused by a huge amount of events generated at instants. Unlike a fixed bandwidth for a frame camera, the bandwidth of an event camera could reach the limits of a vehicle's onboard network capabilities, causing network jamming or package losses. To handle this problem, Khan \textit{et al.}\cite{Khan21} propose an efficient compression algorithm. Another issue is the shadows (in Fig. \ref{fig::event_advantages} (d)), the shadow of a traffic sign on the ground may generate a false alarm.

\section{Conclusion and future works}
Although RGB cameras have cost advantages and are widely applied in vehicles, several inherent limitations impede the development of a better autonomous driving system working in larger ODDs. To overcome these drawbacks, other kinds of sensing modalities are emerging. In this paper, we've reviewed several emerging vision sensors as complementaries for conventional RGB cameras: infrared, range-gated, polarization, and event cameras. Some of them have been already integrated into production cars, such as NIR and LWIR cameras. While some of them are still in early exploration stages, such as the polarization, event and range-gated cameras. With the additional sensing modalities, an autonomous vehicle is expected to be operated in larger ODDs (e.g. at night or through rainy days). Most of the perception algorithms for the introduced sensors are similar to processing RGB images that RGB channels are replaced by infrared or polarization channels. For range-gated and event cameras, since their imaging principles are different, specifically designed algorithms are designed to leverage their unique imaging property. A fusion between those sensors and RGB cameras is more practical in engineering, and several common fusion strategies, i.e. early fusion, half-way fusion and late fusion, are summarized in Fig. \ref{fig::fusion} (b). In the future, we believe that the reviewed sensing technologies will have better maturity, lower price, and more powerful algorithms to utilize their imaging advantages. 

\bibliographystyle{IEEEtran}      
\bibliography{uranus,eventcam_new}   

\begin{thebibliography}{100}
\providecommand{\url}[1]{#1}
\csname url@samestyle\endcsname
\providecommand{\newblock}{\relax}
\providecommand{\bibinfo}[2]{#2}
\providecommand{\BIBentrySTDinterwordspacing}{\spaceskip=0pt\relax}
\providecommand{\BIBentryALTinterwordstretchfactor}{4}
\providecommand{\BIBentryALTinterwordspacing}{\spaceskip=\fontdimen2\font plus
\BIBentryALTinterwordstretchfactor\fontdimen3\font minus
  \fontdimen4\font\relax}
\providecommand{\BIBforeignlanguage}[2]{{%
\expandafter\ifx\csname l@#1\endcsname\relax
\typeout{** WARNING: IEEEtran.bst: No hyphenation pattern has been}%
\typeout{** loaded for the language `#1'. Using the pattern for}%
\typeout{** the default language instead.}%
\else
\language=\csname l@#1\endcsname
\fi
#2}}
\providecommand{\BIBdecl}{\relax}
\BIBdecl

\bibitem{Chris2008}
C.~Urmson \emph{et~al.}, ``{Autonomous driving in urban environments: Boss and
  the Urban Challenge},'' \emph{Journal of Field Robotics}, vol.~25, pp.
  425--466, 2008.

\bibitem{SAELevel}
``{Taxonomy and Definitions for Terms Related to Driving Automation Systems for
  On-Road Motor Vehicles},'' SAE International, Tech. Rep., 2018.

\bibitem{liyouSPM2020}
Y.~Li and J.~Ibanez-Guzman, ``Lidar for autonomous driving: The principles,
  challenges, and trends for automotive lidar and perception systems,''
  \emph{IEEE Signal Processing Magazine}, vol.~37, pp. 50--61, 2020.

\bibitem{radarSPM1027}
S.~M. Patole, M.~Torlak, D.~Wang, and M.~Ali, ``Automotive radars: A review of
  signal processing techniques,'' \emph{IEEE Signal Processing Magazine},
  vol.~34, pp. 22--35, 2017.

\bibitem{Brummelen18}
J.~Van~Brummelen, M.~O’Brien, D.~Gruyer, and H.~Najjaran, ``Autonomous
  vehicle perception: The technology of today and tomorrow,''
  \emph{Transportation Research Part C: Emerging Technologies}, vol.~89, pp.
  384--406, 2018.

\bibitem{Yurtsever20}
E.~Yurtsever, J.~Lambert, A.~Carballo, and K.~Takeda, ``A survey of autonomous
  driving: Common practices and emerging technologies,'' \emph{{IEEE} Access},
  vol.~8, pp. 58\,443--58\,469, 2020.

\bibitem{energy2013}
C.~J. Cleveland and C.~Morris, \emph{Handbook of Energy}.\hskip 1em plus 0.5em
  minus 0.4em\relax Elsevier, 2013.

\bibitem{ASTMG173}
``Standard tables for reference solar spectral irradiances: Direct normal and
  hemispherical on 37° tilted surface,'' American Society for Testing
  Materials, Standard, 2012.

\bibitem{headlamp2003}
J.~D. Boullough \emph{et~al.}, ``An investigation of headlamp glare: Intensity,
  spectrum and size,'' 2003.

\bibitem{lampRegu}
UNECE, ``Regulation of no 112 of the economic commission for europe of the
  united nations (un/ece),'' Tech. Rep., 2014.

\bibitem{duthon2019}
P.~Duthon, M.~Colomb, and F.~Bernardin, ``Light transmission in fog: The
  influence of wavelength on the extinction coefficient,'' \emph{Applied
  Sciences}, vol.~9, pp. 2843--2854, 2019.

\bibitem{photodiode2009}
K.~Murari, R.~Etienne-Cummings, N.~Thakor, and G.~Cauwenberghs, ``Which
  photodiode to use: A comparison of cmos-compatible structures,'' \emph{IEEE
  Sensors Journal}, vol.~9, pp. 752--760, 2009.

\bibitem{Carrasco-Casado2020}
A.~Carrasco-Casado and R.~Mata-Calvo, \emph{Space Optical Links for
  Communication Networks}.\hskip 1em plus 0.5em minus 0.4em\relax Springer
  International Publishing, 2020, pp. 1057--1103.

\bibitem{weikl2020}
K.~Weikl, D.~Schroeder, and W.~Stechele, ``Optimization of automotive color
  filter arrays for traffic light color separation,'' in \emph{Color and
  Imaging Conference}, 2020.

\bibitem{roadSpecular2010}
M.~Roser and P.~Lenz, ``Camera-based bidirectional reflectance measurement for
  road surface reflectivity classification,'' in \emph{IEEE Intelligent
  Vehicles Symposium}, 2010.

\bibitem{duthon2016}
P.~Duthon, F.~Bernardin, F.~Chausse, and M.~Colomb, ``Methodology used to
  evaluate computer vision algorithms in adverse weather conditions,'' in
  \emph{Transportation Research Procedia}, 2016.

\bibitem{irIntech17}
R.~Thakur, \emph{Infrared Sensors for Autonomous Vehicles, in Recent
  Development in Optoelectronic Devices}.\hskip 1em plus 0.5em minus
  0.4em\relax IntechOpen, 2017.

\bibitem{omni2019}
\BIBentryALTinterwordspacing
OmniVision. (2019) Rgb-ir technology. [Online]. Available:
  \url{https://www.ovt.com/purecel-pixel-tech/rgb-ir-technology/faqs}
\BIBentrySTDinterwordspacing

\bibitem{brownCVPR2011}
M.~Brown and S.~Susstrunk, ``Multispectral sift for scene category
  recognition,'' in \emph{IEEE/CVF International Conference on Computer Vision
  and Pattern Recognition (CVPR)}, 2011.

\bibitem{clement2008}
C.~Fredembach and S.~Susstrunk, ``Colouring the near-infrared,'' in
  \emph{Proceeding of 16th Color and Imaging Conference}, 2008.

\bibitem{jesse2016}
J.~R. Dean, ``Using near-infrared photography to better study snow
  microstructure and its variability over time and space,'' Master's thesis,
  Boise State University, 2016.

\bibitem{nyxel2}
\BIBentryALTinterwordspacing
OmniVision. (2019) Nyxel® technology generation2. [Online]. Available:
  \url{https://www.ovt.com/purecel-pixel-tech/nyxel-technology-generation-2}
\BIBentrySTDinterwordspacing

\bibitem{ingaasNIR2012}
E.~de~Borniol \emph{et~al.}, ``High-performance 640 x 512 pixel hybrid ingaas
  image sensor for night vision,'' in \emph{Proc. SPIE 8353, Infrared
  Technology and Applications XXXVIII}, 2012.

\bibitem{chenOE2014}
Z.~Chen, X.~Wang, and R.~Liang, ``Rgb-nir multispectral camera,'' \emph{Optics
  Express}, vol.~22, 2014.

\bibitem{luyue2016}
Y.~M. Lu, C.~Fredembach, M.~Vetterli, and S.~Susstrunk, ``Designing color
  filter arrays for the joint capture of visible and near-infrared images,'' in
  \emph{16th IEEE International Conference on Image Processing (ICIP)}, 2016.

\bibitem{parkSensors2016}
C.~Park and M.~G. Kang, ``Color restoration of rgbn multispectral filter array
  sensor images based on spectral decomposition,'' \emph{Sensors}, vol.~16, no.
  719, 2016.

\bibitem{orit2019}
O.~Skorka, P.~Kane, and R.~Ispasoiu, ``Color correction for rgb sensors with
  dual-band filters for incabin imaging applications,'' in \emph{Electronic
  Imaging, Autonomous Vehicles and Machines Conference}, 2019.

\bibitem{imec2017}
B.~Geelen, N.~Spooren, K.~Tack, A.~Lambrechts, and M.~Jayapala, ``System-level
  analysis and design for rgb-nir cmos camera,'' in \emph{Proc. SPIE 10110,
  Photonic Instrumentation Engineering IV}, 2017.

\bibitem{roleVCSEL}
M.~Dummer, K.~Johnson, S.~Rothwell, K.~Tatah, and M.~Hibbs-Brenner, ``The role
  of vcsels in 3d sensing and lidar,'' in \emph{Proc. SPIE 11692, Optical
  Interconnects XXI}, 2021.

\bibitem{nightglow2013}
R.~H. Vollmerhausen, R.~G. Driggers, and V.~A. Hodgkin, ``Night illumination in
  the near- and short-wave infrared spectral bands and the potential for
  silicon and indium-gallium-arsenide imagers to perform night targeting,''
  \emph{Optical Engineering}, vol.~52, 2013.

\bibitem{ingaasSWIR2013}
F.~Rutz \emph{et~al.}, ``Ingaas infrared detector development for swir imaging
  applications,'' in \emph{Proceedings Volume 8896, Electro-Optical and
  Infrared Systems: Technology and Applications X}, 2013.

\bibitem{SWIR2008}
M.~P. Hansen and D.~S. Malchow, ``Overview of {SWIR} detectors, cameras, and
  applications,'' in \emph{Proc. SPIE 6939, Thermosense XXX}, 2008.

\bibitem{lvCVPR2019}
F.~Lv, Y.~Zheng, B.~Zhang, and F.~Lu, ``{Turn a Silicon Camera into an InGaAs
  Camera},'' in \emph{IEEE/CVF Conference on Computer Vision and Pattern
  Recognition (CVPR)}, 2019.

\bibitem{dai2020}
X.~Dai \emph{et~al.}, ``Tirnet: Object detection in thermal infrared images for
  autonomous driving,'' \emph{Applied Intelligence}, vol.~51, p. 1244–1261,
  2020.

\bibitem{planck2006}
S.~Blundell and K.~Blundell, \emph{Concepts in Thermal Physics}.\hskip 1em plus
  0.5em minus 0.4em\relax Oxford University Press, 2006.

\bibitem{bhan2009}
R.~Bhan \emph{et~al.}, ``Uncooled infrared microbolometer arrays and their
  characterisation techniques,'' \emph{Defence Science Journal}, vol.~59, pp.
  580--589, 2009.

\bibitem{LWIRAbsorber}
J.~Jung \emph{et~al.}, ``Infrared broadband metasurface absorber for reducing
  the thermal mass of a microbolometer,'' \emph{Scientific Reports}, vol.~7,
  no. 430, 2017.

\bibitem{ulis2006}
J.~Tissot \emph{et~al.}, ``Uncooled microbolometer detector: recent
  developments at ulis,'' \emph{Opto-Electronics Review}, vol.~14, pp. 25--32,
  2006.

\bibitem{bololeti}
J.-J. Yon \emph{et~al.}, ``Latest amorphous silicon microbolometer developments
  at leti-lir,'' in \emph{Proc. SPIE 6940, Infrared Technology and Applications
  XXXIV}, 2008.

\bibitem{bolonju}
L.~Yu \emph{et~al.}, ``Low-cost microbolometer type infrared detectors,''
  \emph{Micromachines}, vol.~9, p. 800, 2020.

\bibitem{microbolo2007}
F.~Niklaus \emph{et~al.}, ``Mems-based uncooled infrared bolometer arrays: a
  review,'' in \emph{Proc. SPIE 6836, MEMS/MOEMS Technologies and Applications
  III}, 2007.

\bibitem{mva2014}
R.~Gade and T.~Moeslund, ``Thermal cameras and applications,'' \emph{Machine
  Vision \& Applications}, vol.~25, pp. 245--262, 2014.

\bibitem{janerik2006}
J.-E. Kallhammer, ``Night vision: requirements and possible roadmap for fir and
  nir systems,'' in \emph{Proc. SPIE 6198, Photonics in the Automobile II},
  2006.

\bibitem{tsim2004}
O.~Tsimhoni, J.~Bargman, J.~Minoda, and M.~Flannagan, ``Pedestrian detection
  with near and far infrared night vision enhancement,'' 2004.

\bibitem{omer2006}
O.~Tsimhoni and M.~Flannagan, ``Pedestrian detection with night vision systems
  enhanced by automatic warnings,'' in \emph{Proceedings of the Human Factors
  and Ergonomics Society Annual Meeting}, 2006.

\bibitem{SPIEflir2019}
K.~M. Judd, M.~P. Thornton, and A.~A. Richards, ``Automotive sensing: Assessing
  the impact of fog on lwir, mwir, swir, visible and lidar imaging
  performance,'' in \emph{Proc. SPIE 11002, Infrared Technology and
  Applications XLV}, 2019.

\bibitem{whichSpec2018}
N.~Pinchon \emph{et~al.}, ``All-weather vision for automotive safety: Which
  spectral band?'' in \emph{Advanced Microsystems for Automotive Applications},
  2018.

\bibitem{deville2000}
N.~S. Martinelli and S.~A. Boulanger, ``Cadillac deville thermal imaging night
  vision system,'' in \emph{SAE Technical Paper Series \#2000-01-0323}, 2000.

\bibitem{hoda2002}
T.~Tsuji, H.~Hattori, M.~Watanabe, and N.~Nagaoka, ``Development of
  night-vision system,'' in \emph{IEEE Transactions on Intelligent
  Transportation Systems}, vol.~3, 2002, pp. 203--209.

\bibitem{gazeTracking2015}
F.~Vicente, Z.~Huang, X.~Xiong, F.~Torre, W.~Zhang, and D.~Levi, ``Driver gaze
  tracking and eyes off the road detection system,'' \emph{IEEE Transactions on
  Intelligent Transportation Systems}, vol.~16, pp. 2014--2027, 2015.

\bibitem{gazeIET2016}
L.~Fridman, J.~Lee, B.~Reimer, and T.~Victor, ``Owl and lizard patterns of head
  pose and eye pose in driver gaze classification,'' \emph{IET Computer
  Vision}, vol.~10, pp. 308--314, 2016.

\bibitem{yoon2019}
H.~Yoon \emph{et~al.}, ``Driver gaze detection based on deep residual networks
  using the combined single image of dual near-infrared cameras,'' \emph{IEEE
  Access}, pp. 93\,448 -- 93\,461, 2019.

\bibitem{nhtsa2007}
A.~Eskandarian, R.~Sayed, P.~Delaigue, J.~Blum, and A.~Mortazavi, ``Advanced
  driver fatigue research,'' 2007.

\bibitem{gazeITS2013}
C.~Ahlstrom, K.~Kircher, and A.~Kircher, ``A gaze-based driver distraction
  warning system and its effect on visual behavior,'' \emph{IEEE Transactions
  on Intelligent Transportation Systems}, vol.~14, pp. 965 -- 973, 2013.

\bibitem{dms2019}
C.~Schwarz, J.~Gaspar, T.~Miller, and R.~Yousefian, ``The detection of
  drowsiness using a driver monitoring system,'' \emph{Traffic Injury
  Prevention}, vol.~20, pp. 157--161, 2019.

\bibitem{park2019}
S.~H. Park, H.~S. Yoon, and K.~R. Park, ``Faster r-cnn and geometric
  transformation-based detection of driver’s eyes using multiple
  near-infrared camera sensors,'' \emph{Sensors}, vol.~19, pp. 1--29, 2019.

\bibitem{perclos1998}
D.~Dinges \emph{et~al.}, ``Evaluation of techniques for ocular measurement as
  an index of fatigue and the basis for alertness management,'' 1998.

\bibitem{rti2002}
Q.~Ji and X.~Yang, ``Real-time eye, gaze, and face pose tracking for monitoring
  driver vigilance,'' \emph{Real-Time Imaging}, vol.~8, pp. 357--377, 2002.

\bibitem{iet2009}
M.~J. Flores, J.~M. Armigol, and A.~de~la Escalera, ``Driver drowsiness
  detection system under infrared illumination for an intelligent vehicle,''
  \emph{IET Intelligent Transportation Systems}, vol.~5, no.~4, pp. 241--251,
  2009.

\bibitem{garciaIV2012}
I.~Garcia, S.~Bronte, L.~M. Bergasa, J.~Almazan, and J.~Yebes, ``Vision-based
  drowsiness detector for real driving conditions,'' in \emph{IEEE Intelligent
  Vehicles Symposium}, 2012.

\bibitem{dasguptaITS2018}
A.~Dasgupta, D.~Rahman, and A.~Routray, ``A smartphone-based drowsiness
  detection and warning system for automotive drivers,'' \emph{IEEE
  Transactions on Intelligent Transportation Systems}, vol.~20, no.~11, pp.
  4045 -- 4054, 2018.

\bibitem{ali2018}
R.~A. Naqv, M.~Arsalan, G.~Batchuluun, H.~S. Yoon, and K.~R. Park, ``Deep
  learning-based gaze detection system for automobile drivers using a nir
  camera sensor,'' \emph{Sensors}, vol.~18, 2018.

\bibitem{dongITS2010}
Y.~Dong, Z.~Hu, K.~Uchimura, and N.~Murayama, ``Driver inattention monitoring
  system for intelligent vehicles: A review,'' \emph{IEEE Transactions on
  Intelligent Transportation Systems}, vol.~12, pp. 596--614, 2010.

\bibitem{akinyelu2020}
A.~A. Akinyelu and P.~Blignaut, ``Convolutional neural network-based methods
  for eye gaze estimation: A survey,'' \emph{IEEE Access}, vol.~8, pp.
  142\,581--142\,605, 2020.

\bibitem{algoPPG2017}
W.~Wang, A.~C. den Brinker, S.~Stuijk, and G.~de~Haan, ``Algorithmic principles
  of remote ppg,'' \emph{IEEE Transactions on Biomedical Engineering}, vol.~64,
  pp. 1479--1491, 2017.

\bibitem{ppg2018}
E.~M. Nowara \emph{et~al.}, ``Sparseppg: Towards driver monitoring using
  camera-based vital signs estimation in near-infrared,'' in \emph{IEEE/CVF
  Conference on Computer Vision and Pattern Recognition Workshops (CVPRW)},
  2018.

\bibitem{wangTBE2020}
W.~Wang and A.~C. den Brinker, ``Modified rgb cameras for infrared
  remote-ppg,'' \emph{IEEE Transactions on Biomedical Engineering}, vol.~67,
  pp. 2893 -- 2904, 2020.

\bibitem{tipHR2021}
K.~Kurihara, D.~Sugimura, and T.~Hamamoto, ``Non-contact heart rate estimation
  via adaptive rgb/nir signal fusion,'' \emph{IEEE Transactions on Image
  Processing}, vol.~30, pp. 6528 -- 6543, 2021.

\bibitem{viola2001}
P.~Viola and M.~Jones, ``Rapid object detection using a boosted cascade of
  simple features,'' in \emph{in Proceedings of the IEEE conference on Computer
  Vision and Pattern Recognition}, 2001.

\bibitem{SVMTutorial}
C.~Burges, ``A tutorial on support vector machines for pattern recognition,''
  \emph{Data Mininf and Knowledge Discovery}, vol.~2, pp. 121--167, June 1998.

\bibitem{tvt2004}
Y.~Fang \emph{et~al.}, ``A shape-independent method for pedestrian detection
  with far-infrared images,'' \emph{IEEE Transactions on Vehicular Technology},
  vol.~53, no.~6, pp. 1679--1697, 2004.

\bibitem{xuITS2005}
F.~Xu \emph{et~al.}, ``Pedestrian detection and tracking with night vision,''
  \emph{IEEE Transactions on Intelligent Transportation Systems}, vol.~6, pp.
  63--71, 2005.

\bibitem{autoliv2014}
D.~Forslund and J.~Bjarkerfur, ``Night vision animal detection,'' in \emph{IEEE
  Intelligent Vehicles Symposium}, 2014.

\bibitem{savasturk2015}
D.~Savasturk \emph{et~al.}, ``A comparison study on vehicle detection in far
  infrared and regular images,'' in \emph{IEEE 18th International Conference on
  Intelligent Transportation Systems}, 2015.

\bibitem{chenTIV2019}
Z.~Chen and X.~Huang, ``Pedestrian detection for autonomous vehicle using
  multi-spectral cameras,'' \emph{IEEE Transactions on Intelligent Vehicles},
  vol.~4, pp. 211--219, 2019.

\bibitem{kristo2020}
M.~Kristo \emph{et~al.}, ``Thermal object detection in difficult weather
  conditions using yolo,'' \emph{IEEE Access}, pp. 125\,459--125\,476, 2020.

\bibitem{fasterRCNN}
S.~Ren \emph{et~al.}, ``Faster r-cnn: towards real-time object detection with
  region proposal networks,'' in \emph{Proceedings of the 28th International
  Conference on Neural Information Processing Systems}, 2015.

\bibitem{ssd2016}
W.~Liu \emph{et~al.}, ``{SSD: Single Shot MultiBox Detector},'' in
  \emph{Proceeding of European Conference on Computer Vision}, 2016.

\bibitem{yolov3}
J.~Redmon and A.~Farhadi, ``{YOLOv3: An Incremental Improvement},''
  \emph{arXiv:1804.02767}, 2018.

\bibitem{scut2019}
Z.~Xu \emph{et~al.}, ``Benchmarking a large-scale fir dataset for on-road
  pedestrian detection,'' \emph{Infrared Physics \& Technology}, vol.~96, pp.
  199--208, 2019.

\bibitem{tumas2020}
P.~Tumas \emph{et~al.}, ``Pedestrian detection in severe weather conditions,''
  \emph{IEEE Access}, vol.~8, pp. 62\,775--62\,784, 2020.

\bibitem{osa2021}
R.~Grimming \emph{et~al.}, ``Lwir sensor parameters for deep learning object
  detectors,'' \emph{OSA Continuum}, vol.~4, pp. 529--541, 2021.

\bibitem{fastRCNN}
R.~Girshick, ``Fast r-cnn,'' in \emph{Proceedings of the 2015 IEEE
  International Conference on Computer Vision}, 2015.

\bibitem{fusion2019}
J.~Ma, Y.~Ma, and C.~Li, ``Infrared and visible image fusion methods and
  applications: A survey,'' \emph{Information Fusion}, vol.~45, pp. 153--178,
  2019.

\bibitem{icpr2016}
H.~Choi \emph{et~al.}, ``Multi-spectral pedestrian detection based on
  accumulated object proposal with fully convolutional networks,'' in
  \emph{23rd International Conference on Pattern Recognition (ICPR)}, 2016.

\bibitem{pr2018}
K.~Park, S.~Kim, and K.~Sohn, ``Unified multi-spectral pedestrian detection
  based on probabilistic fusion networks,'' \emph{Pattern Recognition},
  vol.~80, pp. 143--155, 2018.

\bibitem{renaux2020}
G.~Humblot-Renaux \emph{et~al.}, ``Thermal imaging on smart vehicles for person
  and road detection: Can a lazy approach work?'' in \emph{IEEE 23rd
  International Conference on Intelligent Transportation Systems (ITSC)}, 2020.

\bibitem{takumi17}
K.~Takumi \emph{et~al.}, ``Multispectral object detection for autonomous
  vehicles,'' in \emph{Proceedings of the on Thematic Workshops of ACM
  Multimedia}, 2017.

\bibitem{esann2016}
J.~Wagner, V.~Fisher, M.~Herman, and S.~Behnke, ``Multispectral pedestrian
  detection using deep fusion convolutional neural networks,'' in \emph{In
  Proceedings of 24th European Symposium on Artificial Neural Networks,
  Computational Intelligence and Machine Learning (ESANN)}, 2016.

\bibitem{bmvc2016}
J.~Liu \emph{et~al.}, ``Multispectral deep neural networks for pedestrian
  detection,'' in \emph{Proceedings of the British Machine Vision Conference
  (BMVC)}, 2016.

\bibitem{pr2019}
C.~Li \emph{et~al.}, ``Illumination-aware faster r-cnn for robust multispectral
  pedestrian detection,'' \emph{Pattern Recognition}, vol.~85, pp. 161--171,
  2019.

\bibitem{bmvc2018}
------, ``Multispectral pedestrian detection via simultaneous detection and
  segmentation,'' in \emph{In Proceeding of British Machine Vision Conference},
  2018.

\bibitem{IFusion2019}
D.~Guan \emph{et~al.}, ``Fusion of multispectral data through
  illumination-aware deep neural networks for pedestrian detection,''
  \emph{Information Fusion}, vol.~50, pp. 148--157, 2019.

\bibitem{icip2020}
H.~Zhang \emph{et~al.}, ``Multispectral fusion for object detection with cyclic
  fuse-and-refine blocks,'' in \emph{IEEE International Conference on Image
  Processing (ICIP)}, 2020.

\bibitem{yadav2020}
R.~Yadav \emph{et~al.}, ``Cnn based color and thermal image fusion for object
  detection in automated driving,'' in \emph{Irish Machine Vision and Image
  Processing (IMVIP 2020)}, 2020.

\bibitem{cvprw2017}
D.~Konig \emph{et~al.}, ``Fully convolutional region proposal networks for
  multispectral person detection,'' in \emph{IEEE Conference on Computer Vision
  and Pattern Recognition Workshop}, 2017.

\bibitem{zliccv2019}
L.~Zhang \emph{et~al.}, ``Weakly allighed cross-modal learning for
  multispectral pedestrian detection,'' in \emph{IEEE/CVF International
  Conference on Computer Vision (ICCV)}, 2019.

\bibitem{kinjal2021}
K.~Dasgupta \emph{et~al.}, ``Spatio-contextual deep network based multimodal
  pedestrian detection for autonomous driving,'' \emph{arXiv:2105.12713}, 2021.

\bibitem{pelaezIVS2015}
G.~Pelaez \emph{et~al.}, ``Road detection with thermal cameras through 3d
  information,'' in \emph{IEEE Intelligent Vehicles Symposium}, 2015.

\bibitem{YoonIVS2016}
J.~S. Yoon \emph{et~al.}, ``Thermal-infrared based drivable region detection,''
  in \emph{IEEE Intelligent Vehicles Symposium}, 2016.

\bibitem{gated1966}
L.~Gillespie, ``Apparent illuminance as a function of range in gated, laser
  night-viewing systems,'' \emph{Journal of the Optical Society of America},
  vol.~56, pp. 883--887, 1966.

\bibitem{gated2004}
I.~M. Baker \emph{et~al.}, ``A low-noise laser-gated imaging system for
  long-range target identification,'' in \emph{Proc. SPIE 5406, Infrared
  Technology and Applications}, 2004.

\bibitem{marin2020}
A.~M. Pinto and A.~C. Matos, ``Maresye: A hybrid imaging system for underwater
  robotic applications,'' \emph{Information Fusion}, vol.~55, pp. 16--29, 2020.

\bibitem{marioCVPR2020}
M.~Bijelic \emph{et~al.}, ``Seeing through fog without seeing fog: Deep
  multimodal sensor fusion in unseen adverse weather,'' in \emph{Proceedings of
  the IEEE/CVF Conference on Computer Vision and Pattern Recognition}, 2020.

\bibitem{David2006}
O.~David, N.~Kopeika, and B.~Weizer, ``Range gated active night vision system
  for automobiles,'' \emph{Applied Optics}, 2006.

\bibitem{nick2016}
N.~Spooren \emph{et~al.}, ``R{GB-NIR} active gated imaging,'' in
  \emph{Electro-Optical and Infrared Systems: Technology and Applications
  XIII}, 2016.

\bibitem{OE2021}
A.~H. Willitsford \emph{et~al.}, ``Range-gated active short-wave infrared
  imaging for rain penetration,'' \emph{Optical Engineering}, vol.~60, no.~1,
  pp. 1 -- 11, 2021.

\bibitem{apdMatrix2019}
F.~Rutz \emph{et~al.}, ``Ingaas apd matrix sensors for swir gated viewing,''
  \emph{Advanced Optical Technologies}, vol.~8, pp. 445--450, 2019.

\bibitem{spad2014}
S.~Burri \emph{et~al.}, ``Architecture and applications of a high resolution
  gated spad image sensor,'' \emph{Optics Express}, vol.~22, pp.
  17\,573--17\,589, 2014.

\bibitem{optica2020}
K.~Morimoto \emph{et~al.}, ``Megapixel time-gated spad image sensor for 2d and
  3d imaging applications,'' \emph{Optica}, vol.~7, pp. 346--354, 2020.

\bibitem{marioIV2018}
M.~Bijelic, T.~Gruber, and W.~Ritter, ``Benchmarking image sensors under
  adverse weather conditions for autonomous driving,'' in \emph{IEEE
  Intelligent Vehicles Symposium (IV)}, 2018.

\bibitem{gateITSC2020}
S.~Walz, T.~Gruber, W.~Ritter, and K.~Dietmayer, ``Uncertainty depth estimation
  with gated images for 3d reconstruction,'' in \emph{IEEE 23rd International
  Conference on Intelligent Transportation Systems (ITSC)}, 2020.

\bibitem{LEDGated2010}
F.~Christnacher, J.-M. Poyet, M.~Laurenzis, J.-P. Moegline, and F.~Taillade,
  ``Bistatic range-gated active imaging in vehicles with leds or headlights
  illumination,'' in \emph{Proc. SPIE 7675, Photonics in the Transportation
  Industry: Auto to Aerospace III}, 2010.

\bibitem{yoav2014}
Y.~Grauer, ``Active gated imaging in driver assistance system,'' \emph{Advanced
  Optical Technologies}, vol.~3, pp. 151--160, 2014.

\bibitem{yoav2015}
Y.~Grauer and E.~Sonn, ``Active gated imaging for automotive safety
  applications,'' in \emph{Proc. SPIE Video Surveillance and Transportation
  Imaging Applications}, 2015.

\bibitem{gated3D2021}
F.~Julca-Aguilar, J.~Taylor, M.~Bijelic, F.~Mannan, E.~Tseng, and F.~Heide,
  ``Gated3d: Monocular 3d object detection from temporal illumination cues,''
  \emph{arXiv:2102.03602}, 2021.

\bibitem{gated2depth2019}
G.~Tobias, F.~Julca-Aguilar, M.~Bijelic, and F.~Heide, ``Gated2depth: Real-time
  dense lidar from gated images,'' in \emph{The IEEE International Conference
  on Computer Vision (ICCV)}, 2019.

\bibitem{lucid2018}
\BIBentryALTinterwordspacing
{Lucid Vision Labs}. (2018) Beyond conventional imaging: Sony's polarized
  sensor. [Online]. Available:
  \url{https://thinklucid.com/tech-briefs/polarization-explained-sony-polarized-sensor/}
\BIBentrySTDinterwordspacing

\bibitem{wikiPolar}
\BIBentryALTinterwordspacing
{Polarizer: Wikipedia}. (2021) Polarizer. [Online]. Available:
  \url{https://en.wikipedia.org/wiki/Polarizer}
\BIBentrySTDinterwordspacing

\bibitem{polarSN2018}
J.~J. Foster \emph{et~al.}, ``Polarisation vision: overcoming challenges of
  working with a property of light we barely see,'' \emph{The Science of
  Nature}, vol. 105, no.~27, pp. 1--26, 2018.

\bibitem{polarAP1990}
D.~Kliger, \emph{Polarized Light in Optics and Spectroscopy}.\hskip 1em plus
  0.5em minus 0.4em\relax Academic Press, 1990.

\bibitem{hbOptics1996}
M.~Bass, E.~V. Stryland, D.~Williams, and W.~Wolfe, \emph{Handbook of
  Optics}.\hskip 1em plus 0.5em minus 0.4em\relax McGraw-Hill, 1996.

\bibitem{morel2005}
O.~Morel, F.~Meriaudeau, C.~Stolz, and P.~Gorria, ``Polarization imaging
  applied to 3d reconstruction of specular metallic surfaces,'' in \emph{Proc.
  SPIE 5679, Machine Vision Applications in Industrial Inspection XIII}, 2005,
  pp. 178--186.

\bibitem{polarSplit1994}
L.~B. Wolff, ``Polarization camera for computer vision with a beam splitter,''
  \emph{Journal of the Optical Society of America A}, vol.~11, pp. 2935--2945,
  1994.

\bibitem{YouLIHDR}
Y.~Li \emph{et~al.}, ``Multiframe-based high dynamic range monocular vision
  system for advanced driver assistance systems,'' \emph{IEEE Sensors Journal},
  vol.~15, pp. 5433--5441, 2015.

\bibitem{mta2018}
Y.~Wang \emph{et~al.}, ``Efficient road specular reflection removal based on
  gradient properties,'' \emph{Multimedia Tools and Applications}, vol.~77, pp.
  30\,615--30\,631, 2018.

\bibitem{specular2017}
F.~Wang, S.~Ainouz, C.~Petitjean, and A.~Bensrhair, ``Specularity removal: A
  global energy minimization approach based on polarization imaging,''
  \emph{Computer Vision and Image Understanding}, vol. 158, pp. 31--39, 2017.

\bibitem{polarHDR2020}
X.~Wu \emph{et~al.}, ``Hdr reconstruction based on the polarization camera,''
  \emph{IEEE Robotics and Automation Letters}, vol.~5, pp. 5113--5119, 2020.

\bibitem{Posch14}
C.~Posch, T.~Serrano-Gotarredona, B.~Linares-Barranco, and T.~Delbruck,
  ``Retinomorphic event-based vision sensors: Bioinspired cameras with spiking
  output,'' \emph{Proceedings of the {IEEE}}, vol. 102, no.~10, pp. 1470--1484,
  2014.

\bibitem{Gehrig21b}
M.~Gehrig, W.~Aarents, D.~Gehrig, and D.~Scaramuzza, ``{DSEC}: A stereo event
  camera dataset for driving scenarios,'' \emph{{IEEE} Robotic and Automation
  Letters}, 2021.

\bibitem{polarCar2018}
F.~Wang, S.~Ainouz, F.~Meriaudeau, and A.~Bensrhair, ``Polarization-based car
  detection,'' in \emph{25th IEEE International Conference on Image Processing
  (ICIP)}, 2018.

\bibitem{blinITSC2019}
R.~Blin, S.~Ainouz, S.~Canu, and F.~Meriaudeau, ``Road scenes analysis in
  adverse weather conditions by polarization-encoded images and adapted deep
  learning,'' in \emph{IEEE Intelligent Transportation Systems Conference
  (ITSC)}, 2019.

\bibitem{blinCVPRW2020}
------, ``A new multimodal rgb and polarimetric image dataset for road scenes
  analysis,'' in \emph{2020 IEEE/CVF Conference on Computer Vision and Pattern
  Recognition Workshops (CVPRW)}, 2020.

\bibitem{polarLITIS}
------, ``The polarlitis dataset: Road scenes under fog,'' \emph{IEEE
  Transactions on Intelligent Transportation Systems}, p. Early Access, 2021.

\bibitem{visapp2019}
M.~Blanchon \emph{et~al.}, ``Outdoor scenes pixel-wise semantic segmentation
  using polarimetry and fully convolutional network,'' in \emph{14th
  International Joint Conference on Computer Vision, Imaging and Computer
  Graphics Theory and Applications}, 2019.

\bibitem{polarSeg2021}
K.~Xiang, K.~Yang, and K.~Wang, ``Polarization-driven semantic segmentation via
  efficient attention-bridged fusion,'' \emph{Optics Express}, vol.~29, pp.
  4802--4820, 2021.

\bibitem{Gallego20}
G.~Gallego, T.~Delbruck, G.~M. Orchard, C.~Bartolozzi, B.~Taba, A.~Censi,
  S.~Leutenegger, A.~Davison, J.~Conradt, K.~Daniilidis, and D.~Scaramuzza,
  ``Event-based vision: A survey,'' \emph{{IEEE} Transactions on Pattern
  Analysis and Machine Intelligence}, pp. 154--180, 2020.

\bibitem{Chen20b}
G.~Chen and A.~Knoll, ``Event-based neuromorphic vision for autonomous driving:
  A paradigm shift for bio-inspired visual sensing and perception,'' \emph{IEEE
  Signal Processing Magazine}, vol.~37, no.~4, pp. 34--49, 2020.

\bibitem{Finateu20}
T.~Finateu \emph{et~al.}, ``5.10 a 1280×720 back-illuminated stacked temporal
  contrast event-based vision sensor with 4.86µm pixels, 1.066geps readout,
  programmable event-rate controller and compressive data-formatting
  pipeline,'' in \emph{IEEE International Solid- State Circuits Conference},
  2020.

\bibitem{Suh20}
Y.~Suh \emph{et~al.}, ``A 1280×960 dynamic vision sensor with a 4.95-µm pixel
  pitch and motion artifact minimization,'' in \emph{IEEE International
  Symposium on Circuits and Systems}, 2020.

\bibitem{Chen19a}
S.~Chen and M.~Guo, ``Live demonstration: Celex-v: A 1m pixel multi-mode
  event-based sensor,'' in \emph{IEEE/CVF Conference on Computer Vision and
  Pattern Recognition Workshops}, 2019.

\bibitem{Scheerlinck19}
Scheerlinck \emph{et~al.}, ``Ced: Color event camera dataset,'' in
  \emph{IEEE/CVF Conference on Computer Vision and Pattern Recognition
  Workshops}, 2019.

\bibitem{Qiao15}
N.~Qiao, H.~Mostafa, F.~Corradi, M.~Osswald, F.~Stefanini, D.~Sumislawska, and
  G.~Indiveri, ``A reconfigurable on-line learning spiking neuromorphic
  processor comprising 256 neurons and 128k synapses,'' \emph{Frontiers in
  Neuroscience}, vol.~9, p. 141, 2015.

\bibitem{Akopyan15}
F.~Akopyan \emph{et~al.}, ``{TrueNorth}: Design and tool flow of a 65 {mW} 1
  million neuron programmable neurosynaptic chip,'' \emph{{IEEE} Transactions
  on Computer-Aided Design of Integrated Circuits and Systems}, vol.~34,
  no.~10, pp. 1537--1557, 2015.

\bibitem{Davies21}
M.~Davies, A.~Wild, G.~Orchard, Y.~Sandamirskaya, G.~A.~F. Guerra, P.~Joshi,
  P.~Plank, and S.~R. Risbud, ``Advancing neuromorphic computing with loihi: A
  survey of results and outlook,'' \emph{Proceedings of the {IEEE}}, vol. 109,
  no.~5, pp. 911--934, 2021.

\bibitem{Gehrig20b}
D.~Gehrig, H.~Rebecq, G.~Gallego, and D.~Scaramuzza, ``{EKLT}: Asynchronous
  photometric feature tracking using events and frames,'' \emph{International
  Journal of Computer Vision}, vol. 128, no.~3, pp. 601--618, 2020.

\bibitem{Akolkar20}
H.~Akolkar, S.~H. Ieng, and R.~Benosman, ``Real-time high speed motion
  prediction using fast aperture-robust event-driven visual flow,''
  \emph{{IEEE} Transactions on Pattern Analysis and Machine Intelligence}, pp.
  1--1, 2020.

\bibitem{Nunes21}
U.~M. Nunes and Y.~Demiris, ``Robust event-based vision model estimation by
  dispersion minimisation,'' \emph{IEEE Transactions on Pattern Analysis and
  Machine Intelligence}, pp. 1--1, 2021.

\bibitem{Osswald17}
M.~Osswald, S.-H. Ieng, R.~Benosman, and G.~Indiveri, ``A spiking neural
  network model of 3d perception for event-based neuromorphic stereo vision
  systems,'' \emph{Scientific Reports}, vol.~7, no.~1, p. 40703, 2017.

\bibitem{Barbier21}
T.~Barbier, C.~Teuliere, and J.~Triesch, ``Spike timing-based unsupervised
  learning of orientation, disparity, and motion representations in a spiking
  neural network,'' in \emph{{IEEE}/{CVF} Conference on Computer Vision and
  Pattern Recognition Workshops}, 2021.

\bibitem{Viale21}
A.~Viale, A.~Marchisio, M.~Martina, G.~Masera, and M.~Shafique, ``{CarSNN}: An
  efficient spiking neural network for event-based autonomous cars on the loihi
  neuromorphic research processor,'' in \emph{International Joint Conference on
  Neural Networks}, 2021.

\bibitem{Paredes20}
F.~Paredes-Vallés, K.~Y.~W. Scheper, and G.~C. H. E.~d. Croon, ``Unsupervised
  learning of a hierarchical spiking neural network for optical flow
  estimation: From events to global motion perception,'' \emph{{IEEE}
  Transactions on Pattern Analysis and Machine Intelligence}, vol.~42, no.~8,
  pp. 2051--2064, 2020.

\bibitem{Parameshwara21a}
C.~M. Parameshwara, S.~Li, C.~Fermüller, N.~J. Sanket, M.~S. Evanusa, and
  Y.~Aloimonos, ``{SpikeMS}: Deep spiking neural network for motion
  segmentation,'' \emph{{arXiv}:2105.06562}, 2021.

\bibitem{Kreiser18a}
R.~Kreiser, ``A neuromorphic approach to path integration: A head-direction
  spiking neural network with vision-driven reset,'' in \emph{{IEEE}
  International Symposium on Circuits and Systems}, 2018.

\bibitem{Kreiser20}
R.~Kreiser \emph{et~al.}, ``Error estimation and correction in a spiking neural
  network for map formation in neuromorphic hardware,'' in \emph{{IEEE}
  International Conference on Robotics and Automation}, 2020.

\bibitem{Stagsted20a}
R.~Stagsted, A.~Vitale, J.~Binz, A.~Renner, L.~Bonde~Larsen, and
  Y.~Sandamirskaya, ``Towards neuromorphic control: A spiking neural network
  based {PID} controller for {UAV},'' in \emph{Robotics: Science and Systems
  {XVI}}, 2020.

\bibitem{Vitale21}
A.~Vitale, A.~Renner, C.~Nauer, D.~Scaramuzza, and Y.~Sandamirskaya,
  ``Event-driven vision and control for {UAVs} on a neuromorphic chip,''
  \emph{{arXiv}:2108.03694}, 2021.

\bibitem{Ruckauer19}
B.~Rückauer, N.~Känzig, S.-C. Liu, T.~Delbruck, and Y.~Sandamirskaya,
  ``Closing the accuracy gap in an event-based visual recognition task,''
  \emph{{arXiv}:1906.08859 [cs]}, 2019.

\bibitem{Salvatore20}
N.~Salvatore, S.~Mian, C.~Abidi, and A.~D. George, ``A neuro-inspired approach
  to intelligent collision avoidance and navigation,'' in \emph{{AIAA}/{IEEE}
  39th Digital Avionics Systems Conference}, 2020.

\bibitem{Zou17}
D.~Zou \emph{et~al.}, ``Robust dense depth maps generations from sparse {DVS}
  stereos,'' in \emph{British Machine Vision Conference}, 2017.

\bibitem{Rebecq17b}
H.~Rebecq, T.~Horstschaefer, and D.~Scaramuzza, ``Real-time visual-inertial
  odometry for event cameras using keyframe-based nonlinear optimization,'' in
  \emph{British Machine Vision Conference}, 2017.

\bibitem{Joubert19}
D.~Joubert, M.~Hébert, H.~Konik, and C.~Lavergne, ``Characterization setup for
  event-based imagers applied to modulated light signal detection,''
  \emph{Applied Optics}, vol.~58, no.~6, pp. 1305--1317, 2019.

\bibitem{Akolkar15}
H.~Akolkar \emph{et~al.}, ``What can neuromorphic event-driven precise timing
  add to spike-based pattern recognition?'' \emph{Neural Computation}, vol.~27,
  no.~3, pp. 561--593, 2015.

\bibitem{Moeys16}
D.~P. Moeys \emph{et~al.}, ``Steering a predator robot using a mixed
  frame/event-driven convolutional neural network,'' in \emph{Second
  International Conference on Event-based Control, Communication, and Signal
  Processing}, 2016.

\bibitem{Moeys18}
------, ``{PRED}18: Dataset and further experiments with {DAVIS} event camera
  in predator-prey robot chasing,'' \emph{arXiv:1807.03128}, 2018.

\bibitem{Afshar19}
S.~Afshar, T.~J. Hamilton, J.~Tapson, A.~van Schaik, and G.~Cohen,
  ``Investigation of event-based surfaces for high-speed detection,
  unsupervised feature extraction, and object recognition,'' \emph{Frontiers in
  Neuroscience}, vol.~12, p. 1047, 2019.

\bibitem{Paredes21}
F.~Paredes-Vallés, J.~Hagenaars, and G.~de~Croon, ``Self-supervised learning
  of event-based optical flow with spiking neural networks,''
  \emph{{arXiv}:2106.01862}, 2021.

\bibitem{Chen18}
N.~F.~Y. Chen, ``Pseudo-labels for supervised learning on dynamic vision sensor
  data, applied to object detection under ego-motion,'' in \emph{{IEEE}/{CVF}
  Conference on Computer Vision and Pattern Recognition Workshops}, 2018.

\bibitem{Chen20a}
G.~Chen \emph{et~al.}, ``{NeuroIV}: Neuromorphic vision meets intelligent
  vehicle towards safe driving with a new database and baseline evaluations,''
  \emph{{IEEE} Transactions on Intelligent Transportation Systems}, pp. 1--13,
  2020.

\bibitem{Perot20}
E.~Perot, P.~de~Tournemire, D.~Nitti, J.~Masci, and A.~Sironi, ``Learning to
  detect objects with a 1 megapixel event camera,'' \emph{Advances in Neural
  Information Processing Systems}, vol.~33, pp. 16\,639--16\,652, 2020.

\bibitem{Zhu19a}
A.~Z. Zhu \emph{et~al.}, ``Unsupervised event-based learning of optical flow,
  depth, and egomotion,'' in \emph{Proceedings of the IEEE/CVF Conference on
  Computer Vision and Pattern Recognition}, 2019.

\bibitem{Tulyakov19}
S.~Tulyakov, F.~Fleuret, M.~Kiefel, P.~Gehler, and M.~Hirsch, ``Learning an
  event sequence embedding for dense event-based deep stereo,'' in
  \emph{{IEEE}/{CVF} International Conference on Computer Vision}, 2019.

\bibitem{Cannici20}
M.~Cannici, M.~Ciccone, A.~Romanoni, and M.~Matteucci, ``A differentiable
  recurrent surface for asynchronous event-based data,'' in \emph{European
  Conference on Computer Vision}, 2020.

\bibitem{Li19}
J.~Li, S.~Dong, Z.~Yu, Y.~Tian, and T.~Huang, ``Event-based vision enhanced: A
  joint detection framework in autonomous driving,'' in \emph{{IEEE}
  International Conference on Multimedia and Expo}, 2019.

\bibitem{Zhu18a}
A.~Z. Zhu, D.~Thakur, T.~Özaslan, B.~Pfrommer, V.~Kumar, and K.~Daniilidis,
  ``The multivehicle stereo event camera dataset: An event camera dataset for
  3d perception,'' \emph{{IEEE} Robotics and Automation Letters}, vol.~3,
  no.~3, pp. 2032--2039, 2018.

\bibitem{Binas17}
J.~Binas, D.~Niel, S.-C. Liu, and T.~Delbruck, ``{DDD}17: End-to-end {DAVIS}
  driving dataset,'' \emph{Workshop on Machine Learning for Autonomous
  Vehicles}, 2017.

\bibitem{Cheng19}
W.~Cheng \emph{et~al.}, ``{DET}: A high-resolution {DVS} dataset for lane
  extraction,'' in \emph{{IEEE}/{CVF} Conference on Computer Vision and Pattern
  Recognition Workshops}, 2019.

\bibitem{Cao21}
H.~Cao \emph{et~al.}, ``Fusion-based feature attention gate component for
  vehicle detection based on event camera,'' \emph{{IEEE} Sensors Journal},
  vol.~21, no.~21, pp. 24\,540--24\,548, 2021.

\bibitem{Hu20b}
Y.~Hu, T.~Delbruck, and S.-C. Liu, ``Learning to exploit multiple vision
  modalities by using grafted networks,'' in \emph{European Conference on
  Computer Vision}, 2020.

\bibitem{Hu21}
Y.~Hu, S.-C. Liu, and T.~Delbruck, ``v2e: From video frames to realistic dvs
  events,'' in \emph{IEEE/CVF Conference on Computer Vision and Pattern
  Recognition Workshops}, June 2021.

\bibitem{Chen19b}
G.~Chen \emph{et~al.}, ``Multi-cue event information fusion for pedestrian
  detection with neuromorphic vision sensors,'' \emph{Frontiers in
  Neurorobotics}, vol.~13, pp. 10--16, 2019.

\bibitem{Jiang19}
Z.~Jiang \emph{et~al.}, ``Mixed frame-/event-driven fast pedestrian
  detection,'' in \emph{International Conference on Robotics and Automation},
  2019.

\bibitem{Cladera20}
F.~C. Ojeda, A.~Bisulco, D.~Kepple, V.~Isler, and D.~D. Lee, ``On-device event
  filtering with binary neural networks for pedestrian detection using
  neuromorphic vision sensors,'' in \emph{{IEEE} International Conference on
  Image Processing}, 2020.

\bibitem{Wan21}
J.~Wan \emph{et~al.}, ``Event-based pedestrian detection using dynamic vision
  sensors,'' \emph{Electronics}, vol.~10, no.~8, p. 888, 2021.

\bibitem{Mitrokhin18}
A.~Mitrokhin, C.~Fermüller, C.~Parameshwara, and Y.~Aloimonos, ``Event-based
  moving object detection and tracking,'' in \emph{{IEEE}/{RSJ} International
  Conference on Intelligent Robots and Systems}, 2018.

\bibitem{Stoffregen19a}
T.~Stoffregen \emph{et~al.}, ``Event-based motion segmentation by motion
  compensation,'' in \emph{{IEEE}/{CVF} International Conference on Computer
  Vision}, 2019.

\bibitem{Zhou21a}
Y.~Zhou, G.~Gallego, X.~Lu, S.~Liu, and S.~Shen, ``Event-based motion
  segmentation with spatio-temporal graph cuts,'' \emph{{IEEE} Transactions on
  Neural Networks and Learning Systems}, pp. 1--13, 2021.

\bibitem{Parameshwara21b}
C.~M. Parameshwara \emph{et~al.}, ``0-{MMS}: Zero-shot multi-motion
  segmentation with a monocular event camera,'' in \emph{{IEEE} International
  Conference on Robotics and Automation}, 2021.

\bibitem{Mondal21}
A.~Mondal \emph{et~al.}, ``Moving object detection for event-based vision using
  graph spectral clustering,'' in \emph{{IEEE}/{CVF} International Conference
  on Computer Vision Workshops}, 2021.

\bibitem{Chen20c}
G.~Chen \emph{et~al.}, ``{EDDD}: Event-based drowsiness driving detection
  through facial motion analysis with neuromorphic vision sensor,''
  \emph{{IEEE} Sensors Journal}, vol.~20, no.~11, pp. 6170--6181, 2020.

\bibitem{FLIRADAS}
\BIBentryALTinterwordspacing
FLIR. (2019) {FLIR} thermal sensing for {ADAS}. [Online]. Available:
  \url{https://www.flir.com/oem/adas/adas-dataset-form/}
\BIBentrySTDinterwordspacing

\bibitem{kaistDS2018}
Y.~Choi \emph{et~al.}, ``Kaist multi-spectral day/night data set for autonomous
  and assisted driving,'' \emph{IEEE Transactions on Intelligent Transportation
  Systems}, vol.~19, pp. 934--948, 2018.

\bibitem{ranus2018}
G.~Choe \emph{et~al.}, ``Ranus: Rgb and nir urban scene dataset for deep scene
  parsing,'' \emph{IEEE Robotics and Automation Letters}, vol.~3, pp. 1808 --
  1815, 2018.

\bibitem{denseDataset}
\BIBentryALTinterwordspacing
DENSE. (2019) Dense dataset. [Online]. Available:
  \url{https://www.uni-ulm.de/en/in/driveu/projects/dense-datasets/}
\BIBentrySTDinterwordspacing

\bibitem{Hu20a}
Y.~Hu, J.~Binas, D.~Neil, S.-C. Liu, and T.~Delbruck, ``{DDD}20 end-to-end
  event camera driving dataset: Fusing frames and events with deep learning for
  improved steering prediction,'' in \emph{{IEEE} 23rd International Conference
  on Intelligent Transportation Systems}, 2020.

\bibitem{Alonso19}
I.~Alonso and A.~C. Murillo, ``{EV}-{SegNet}: Semantic segmentation for
  event-based cameras,'' in \emph{Proceedings of the {IEEE}/{CVF} Conference on
  Computer Vision and Pattern Recognition Workshops}, 2019.

\bibitem{Sironi18}
A.~Sironi, M.~Brambilla, N.~Bourdis, X.~Lagorce, and R.~Benosman, ``{HATS}:
  Histograms of averaged time surfaces for robust event-based object
  classification,'' in \emph{{IEEE}/{CVF} Conference on Computer Vision and
  Pattern Recognition}, 2018.

\bibitem{Zhu18b}
A.~Z. Zhu, L.~Yuan, K.~Chaney, and K.~Daniilidis, ``{EV}-{FlowNet}:
  Self-supervised optical flow estimation for event-based cameras,''
  \emph{Robotics: Science and Systems {XIV}}, 2018.

\bibitem{Hu19}
Y.~Hu \emph{et~al.}, ``Slasher: Stadium racer car for event camera end-to-end
  learning autonomous driving experiments,'' in \emph{{IEEE} International
  Conference on Artificial Intelligence Circuits and Systems}, 2019.

\bibitem{Rebecq19}
H.~Rebecq, R.~Ranftl, V.~Koltun, and D.~Scaramuzza, ``High speed and high
  dynamic range video with an event camera,'' \emph{{IEEE} Transactions on
  Pattern Analysis and Machine Intelligence}, pp. 1--1, 2019.

\bibitem{Miao19}
S.~Miao, G.~Chen, X.~Ning, Y.~Zi, K.~Ren, Z.~Bing, and A.~Knoll, ``Neuromorphic
  vision datasets for pedestrian detection, action recognition, and fall
  detection,'' \emph{Frontiers in Neurorobotics}, vol.~13, 2019.

\bibitem{Tournemire20}
P.~de~Tournemire, ``A large scale event-based detection dataset for
  automotive,'' \emph{{arXiv}:2001.08499}, 2020.

\bibitem{Fischer20}
T.~Fischer and M.~Milford, ``Event-based visual place recognition with
  ensembles of temporal windows,'' \emph{{IEEE} Robotics and Automation
  Letters}, vol.~5, no.~4, pp. 6924--6931, 2020.

\bibitem{Hidalgo20}
D.~G. Javier Hidalgo-Carrio and D.~Scaramuzza, ``Learning monocular dense depth
  from events,'' \emph{{IEEE} International Conference on 3D Vision}, 2020.

\bibitem{Gehrig21c}
M.~Gehrig, M.~Millh\"ausler, D.~Gehrig, and D.~Scaramuzza, ``E-raft: Dense
  optical flow from event cameras,'' in \emph{International Conference on 3D
  Vision}, 2021.

\bibitem{Gehrig21a}
D.~Gehrig, M.~Rüegg, M.~Gehrig, J.~Hidalgo-Carrio, and D.~Scaramuzza,
  ``Combining events and frames using recurrent asynchronous multimodal
  networks for monocular depth prediction,'' \emph{{IEEE} Robotic and
  Automation Letters}, 2021.

\bibitem{Delbruck21}
T.~Delbruck, R.~Graca, and M.~Paluch, ``Feedback control of event cameras,'' in
  \emph{IEEE/CVF Conference on Computer Vision and Pattern Recognition
  Workshops}, 2021.

\bibitem{Khan21}
N.~Khan, K.~Iqbal, and M.~G. Martini, ``Time-aggregation-based lossless video
  encoding for neuromorphic vision sensor data,'' \emph{IEEE Internet of Things
  Journal}, vol.~8, no.~1, pp. 596--609, 2021.

\end{thebibliography}

\end{document}